\documentclass[11pt]{article}

\usepackage[final]{acl}

\usepackage{times}
\usepackage{latexsym}

\usepackage[T1]{fontenc}
\usepackage[utf8]{inputenc}

\usepackage{microtype}

\usepackage{inconsolata}

\usepackage{graphicx}

\usepackage{microtype}
\usepackage{tcolorbox}
\usepackage{colortbl} % 为表格增加颜色
\usepackage{newfloat}
\usepackage{listings}
\usepackage{inconsolata}
\usepackage{graphicx}
\usepackage{booktabs}
\usepackage{wrapfig}
\usepackage{enumitem}
\usepackage{subfigure}
\usepackage{multirow}
\usepackage{algorithm}
\usepackage{algpseudocode}
\usepackage{algorithmicx}

\definecolor{R1}{HTML}{56C596}         % 绿色
\definecolor{R1-Searcher}{HTML}{4DA5D9} % 浅蓝
\definecolor{Search-R1}{HTML}{3C88C6}   % 蓝色
\definecolor{Graph-R1}{HTML}{6A0DAD}   

\newcommand{\think}[1]{\textcolor{blue}{\texttt{<think>}} #1 \textcolor{blue}{\texttt{</think>}}}
\newcommand{\search}[1]{\textcolor{orange}{\texttt{<query>}} #1 \textcolor{orange}{\texttt{</query>}}}
\newcommand{\info}[1]{\textcolor{teal}{\texttt{<knowledge>}} #1 \textcolor{teal}{\texttt{</knowledge>}}}
\newcommand{\answer}[1]{\textcolor{purple}{\texttt{<answer>}} #1 \textcolor{purple}{\texttt{</answer>}}}
\usepackage{amsmath}
\usepackage{amsfonts}
\title{AdaptR1: Reinforcement Learning Based Adaptive Interleaved Thinking in Multi-hop Question Answering}

\author{
Yuxin Wang$^{1,2}$\!,\textsuperscript{*}
Jiahao Lu$^{1,3}$\!,\textsuperscript{*} 
Qifeng Wu$^{1}$\!,\textsuperscript{*}
Shicheng Fang$^{1,3}$, \\ 
\textbf{Chuanyuan Tan$^{4}$}, 
\textbf{Yining Zheng$^{1}$}, 
\textbf{Xuanjing Huang$^{1,2}$}, 
\textbf{Xipeng Qiu$^{1,3}$} \\
$^{1}$Computer Science, Fudan University \\
$^{2}$Institute of Modern Languages and Linguistics, Fudan University \\
$^{3}$Shanghai Innovation Institute \\
$^{4}$Soochow University \\
\texttt{\{wangyuxin21, 25113050083, 25213050409, 25113050022\}@m.fudan.edu.cn} \\
\texttt{\{ynzheng19, xjhuang, xpqiu\}@fudan.edu.cn} \\
\texttt{cytan17726@stu.suda.edu.cn} \\
}

\begin{document}

\maketitle
\makeatletter
\ifacl@anonymize\else
\renewcommand{\thefootnote}{}
\footnotetext{*Equal contribution.}
\renewcommand{\thefootnote}{\arabic{footnote}}
\fi
\makeatother

\begin{abstract}
Large Language Models (LLMs) have achieved remarkable performance in complex reasoning tasks through Chain-of-Thought (CoT) prompting. However, this approach often leads to ``over-thinking,'' where models generate unnecessarily long reasoning traces for simple queries and incur avoidable inference cost. While recent work has explored adaptive reasoning, existing methods typically make a single query-level decision about whether to reason. This overlooks the dynamic nature of multi-step tasks, where the need for explicit reasoning varies across intermediate stages. To address this limitation, we introduce AdaptR1, a Reinforcement Learning (RL) based framework for adaptive interleaved thinking in multi-hop Question Answering (QA). Unlike previous approaches that require Supervised Fine-Tuning (SFT) for cold-start initialization, AdaptR1 uses a fully RL-based strategy with a quality-gated efficiency reward to dynamically allocate reasoning budgets at each step. Under the Graph-R1 setting, AdaptR1 reduces average think tokens by 69.71\%, with a 90.35\% reduction on HotpotQA, while maintaining performance comparable to or better than standard baselines. Furthermore, our analysis reveals that overthinking in multi-hop reasoning is not uniformly distributed but occurs predominantly during the initial planning stages, highlighting the effectiveness of step-wise adaptive budget allocation.
\end{abstract}

\section{Introduction}

\begin{figure}[htbp]
    \centering
    \includegraphics[width=\linewidth]{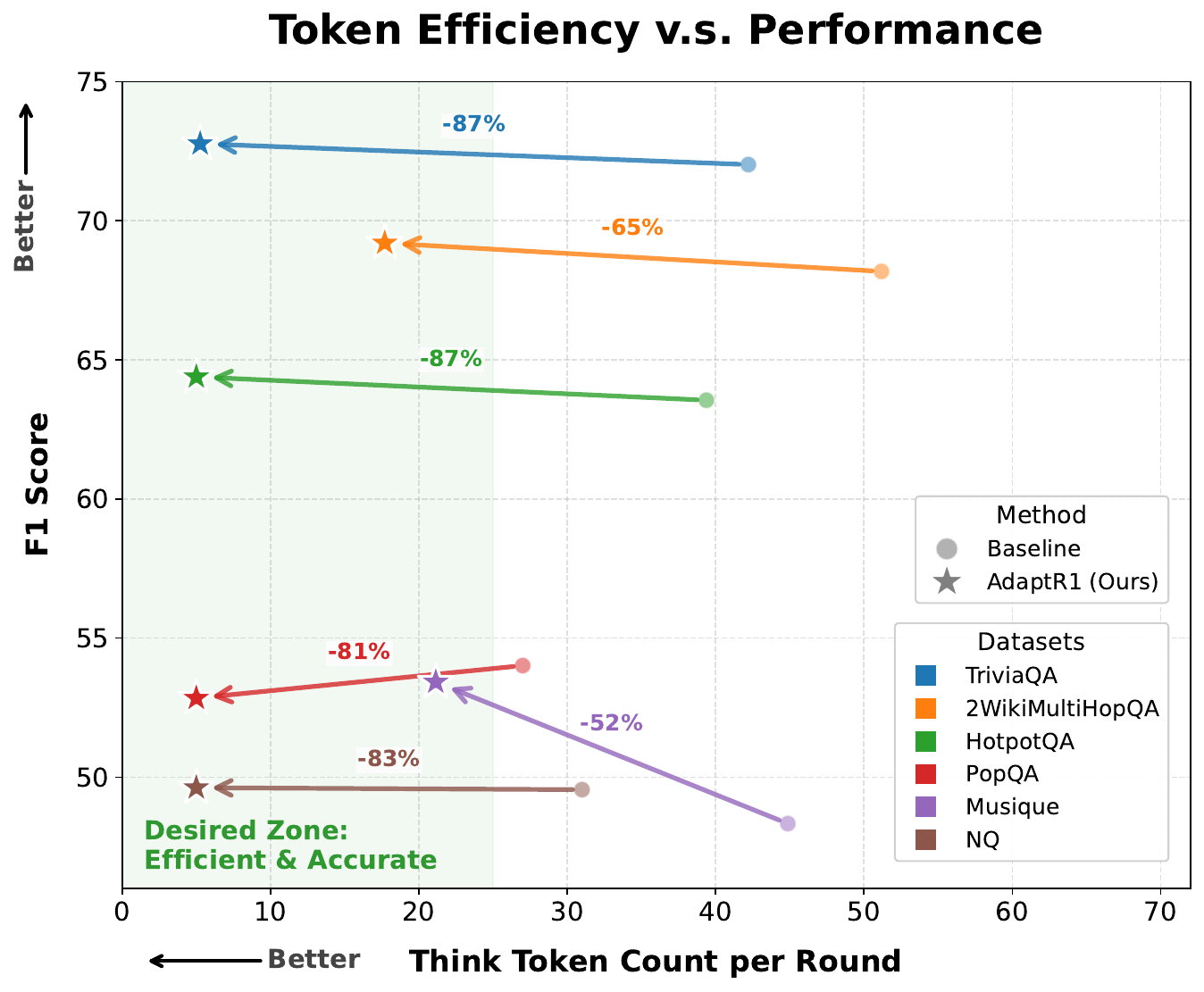}
    \caption{Comparison of F1 scores across RAG benchmarks. AdaptR1 maintains comparable performance while reducing think tokens.}
    \label{fig:adaptr1first}
\end{figure}
Recently, large language models (LLMs)\citep{GPT-4o,GPT-o1} have demonstrated remarkable capabilities across a wide range of natural language understanding and generation tasks. Despite these capabilities, LLMs still struggle with tasks that require complex and multi-step reasoning, such as mathematical problem solving, logical inference, and planning. To address this limitation, researchers have explored methods to elicit stronger reasoning behavior. Relevant methods\citep{GPT-o1,GRPO} mainly include prompt-based techniques, such as Chain-of-Thought (CoT)\citep{wei2022chain} prompting, which encourages models to generate intermediate reasoning steps, and training-based approaches, including Supervised Fine-tuning (SFT)\citep{SFT} with reasoning traces and Reinforcement Learning (RL)\citep{GRPO}, which explicitly incorporate reasoning processes into the model.

\begin{figure*}[htbp]
    \centering
    
    \includegraphics[width=\linewidth]{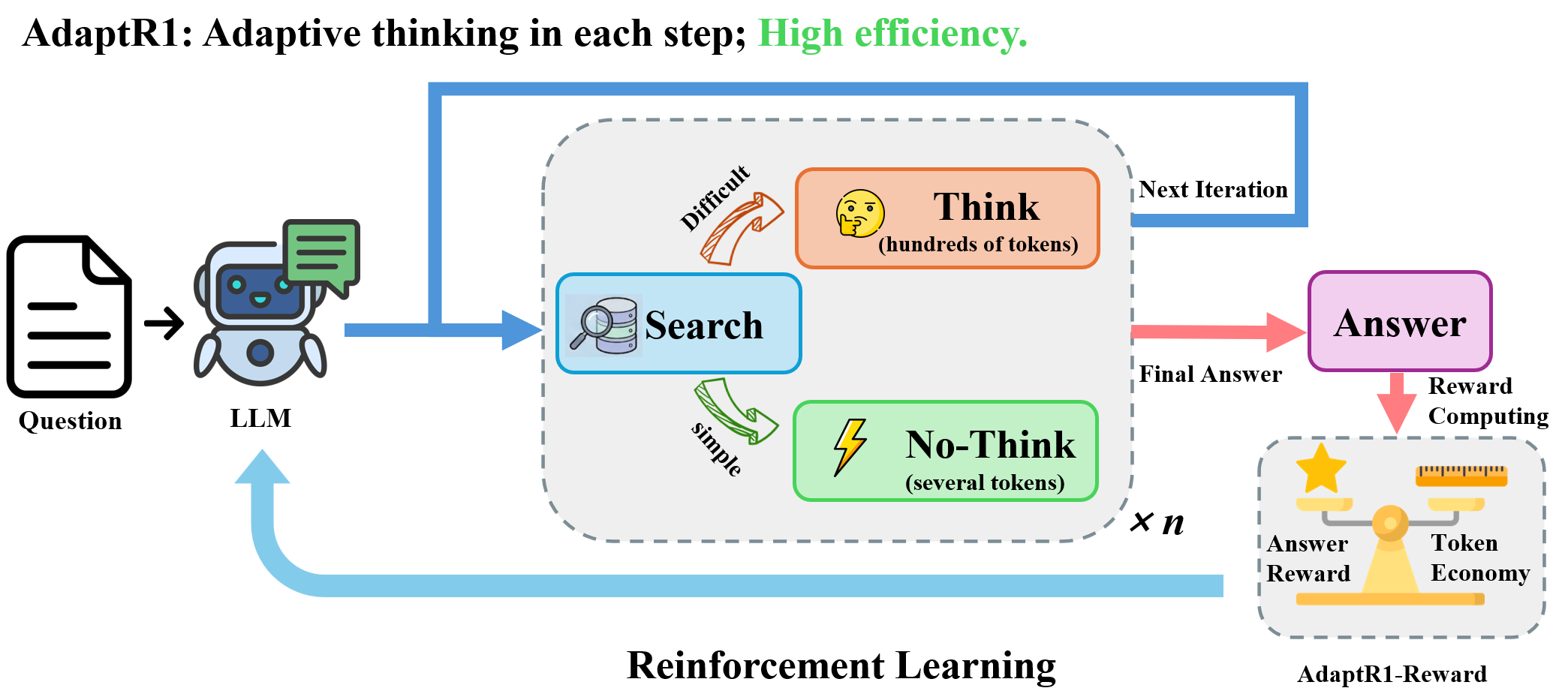}
    \caption{Framework of AdaptR1. RL teaches the model to skip explicit thinking at selected intermediate steps, reducing over-thinking and token usage.}
    \label{fig:AdaptR1-main}
\end{figure*}
However, the introduction of CoT has also led to an emerging issue: overthinking\citep{kumar2025overthink,sui2025stop,nayab2024concise}. Instead of allocating reasoning effort proportionally to task difficulty, CoT often induces LLMs to produce unnecessarily long reasoning traces even for simple queries, increasing both inference time and computational cost. To mitigate this, several research directions have emerged. For instance, SFT with preferred reasoning lengths\citep{chen2025overthink, shen2025dast, luo2025o1pruner} or reinforcement learning with length-based penalties\citep{arora2025training, kimiteam2025kimik15} encourages concise reasoning traces, while adaptive reasoning or selective thinking techniques\citep{zhang2025adaptthinkreasoningmodelslearn,tu2025learningthinkshapingadaptive,ma2025reasoning,chen2025a2fmadaptiveagentfoundation} attempt to adjust the depth of reasoning based on query complexity.

While existing methods aim to balance reasoning depth and computational efficiency, current adaptive thinking strategies typically make a single global decision---whether to think or not---on a per-query basis. However, no prior work investigates adaptive \textbf{interleaved} thinking in multi-step reasoning settings, where decisions to reason, skip, or adjust thinking effort occur dynamically across intermediate steps.

This raises a key question: \textit{Is it necessary for an LLM to think at every step of multi-hop reasoning? If over-thinking exists within intermediate steps, how can an LLM learn to select \textit{Think} or \textit{No-Think} adaptively based on the difficulty of each step?} 

In this paper, we introduce \textbf{AdaptR1}, an RL-based adaptive interleaved thinking method for multi-hop question answering. AdaptR1 enables adaptive reasoning at each stage of a multi-step process, allowing models to allocate reasoning effort more efficiently. Motivated by AdaptThink\citep{zhang2025adaptthinkreasoningmodelslearn}, we use \texttt{<think>no\_think</think>} to denote skipping explicit thinking in the current step. Under the Graph-R1 setting, our method reduces average think tokens by 69.71\% while maintaining answer quality. Our analysis characterizes where efficient thinking emerges across multi-hop reasoning steps and how overthinking appears within the process. Our contributions are as follows:
\begin{itemize}
    \item We study adaptive interleaved thinking for multi-hop QA and show that over-thinking appears in intermediate reasoning steps and can be reduced through learning.
    \item We propose an RL-only adaptive thinking method that avoids SFT cold-start trajectories, and we design a quality-gated efficiency reward for QA tasks with continuous answer rewards such as F1.
    \item AdaptR1 achieves comparable or better performance in multi-hop question answering with a 69.71\% average think-token reduction under the Graph-R1 setting. Extensive analysis shows that over-thinking is concentrated in the early stages of multi-hop reasoning rather than the final synthesis stage.
    
\end{itemize}

\section{Related Works}
\textbf{Efficient Reasoning in LRMs.} Following recent observations regarding the ``over-thinking'' phenomenon in long Chain-of-Thought (CoT) reasoning, adaptive thinking strategies have garnered significant attention. Existing approaches to efficiency generally fall into two categories. The first involves intrinsic model modifications, achieved either through integrating length-based rewards in reinforcement learning (RL)\cite{arora2025training, kimiteam2025kimik15,aggarwal2025l1,hou2025thinkprunepruninglongchainofthought,lou2025adacotparetooptimaladaptivechainofthought}, supervised fine-tuning (SFT) on concise responses\cite{chen2025overthink, shen2025dast, luo2025o1pruner,ma2025cotvalvelengthcompressiblechainofthoughttuning,kang2025c3ot}, or by amalgamating reasoning and non-reasoning parameters~\cite{wu2025unlocking}. The second category empowers LLMs to adaptively modulate their reasoning process based on query complexity. Prominent examples include AdaptThink~\citep{zhang2025adaptthinkreasoningmodelslearn}, AutoThink~\citep{tu2025learningthinkshapingadaptive}, HiPO~\citep{deng2025hipohybridpolicyoptimization}, and ARM~\citep{wu2025arm}; notably, A$^2$FM~\citep{chen2025a2fmadaptiveagentfoundation} extends this framework to encompass instant, reasoning, and agentic modes. Our research aligns with this second paradigm. However, these methods usually make a single query-level routing decision, whereas interleaved multi-hop QA requires repeated decisions after each retrieval result. This step-wise setting also makes direct comparison with single-turn pruning methods less informative, because they do not support the reason-search-answer loop evaluated here. AdaptR1 therefore addresses the unexplored challenge of mitigating over-thinking within the granular steps of multi-hop reasoning, and it learns this behavior directly through RL without SFT cold-start trajectories.

\textbf{Multi-hop Question Answering.} Methodologies for multi-hop Question Answering (QA) can be broadly classified into training-free and training-based paradigms. Training-free methods employ prompting strategies such as Chain-of-Thought (CoT)\citep{wei2022chain} and various retrieval-augmented frameworks including IRCoT\citep{trivedi2022interleaving}, ITER-RETGEN~\citep{shao2023enhancing}, WebGPT~\citep{nakano2021webgpt}, ReAct~\citep{yao2023react}, Self-RAG~\citep{asai2024self}, Self-ask~\citep{press2023measuringnarrowingcompositionalitygap}, and FLARE~\citep{jiang2023active}. Conversely, training-based methods---such as R1-Searcher~\citep{R1-Searcher}, DeepResearcher~\citep{zheng-etal-2025-deepresearcher}, R3-RAG~\citep{r3-rag}, DeepRAG~\citep{DeepRAG}, Search-R1~\citep{Search-R1}, and Graph-R1~\citep{Graph-R1}---utilize SFT or RL to cultivate step-by-step reasoning capabilities. Despite these advancements, current literature lacks a mechanism to address over-thinking specifically within multi-hop QA contexts. Our proposed method fills this gap by implementing an RL-exclusive adaptive thinking strategy tailored for these scenarios.

\section{Preliminaries}

AdaptR1 adds one adaptive instruction to the native parent prompt: at each scheduled reasoning slot, the model can either generate explicit reasoning inside \texttt{<think>}...\texttt{</think>} or emit \texttt{<think>no\_think</think>} to skip it. In the Graph-R1 setting, tool calls use \texttt{<query>}...\texttt{</query>} and retrieved evidence is returned inside \texttt{<knowledge>}...\texttt{</knowledge>}; Search-AdaptR1 preserves Search-R1's native \texttt{<search>}/\texttt{<information>} interface. The full Graph-AdaptR1 prompt is provided in Appendix~\ref{appendix:prompt}.

\textbf{GRPO.} Group Relative Policy Optimization (GRPO)\citep{GRPO} samples grouped rollouts for each question and updates the policy with group-normalized sequence-level advantages. In AdaptR1, the standard sequence reward is replaced by the adaptive reward in Eq.~\ref{eq:AdaptR1reward}; the full GRPO objective is provided in Appendix~\ref{sec:appendix-grpo}.

\noindent \textbf{Interleaved Thinking.} R1-like methods employ an iterative process of reasoning and retrieval to synthesize a final output. This process is modeled as an action sequence $\mathcal{A} = [a_0, a_1, \dots, a_t]$, initialized with $a_0 = \textbf{Think}$. For subsequent steps $i > 0$, the transition logic dictates that if the preceding action $a_{i-1}$ was a reasoning step (\textbf{Think}), the subsequent action $a_i$ must be either \textbf{Search} or \textbf{Answer}. Conversely, if $a_{i-1} \neq \textbf{Think}$, the system defaults to \textbf{Think}. The action space is defined as follows:
\begin{itemize}
    \item \textbf{Think}: Derives reasoning steps utilizing existing internal knowledge.
    \item \textbf{Search}: Queries an external knowledge base to retrieve supplementary information.
    \item \textbf{Answer}: Terminates the sequence by providing the final response once information sufficiency is achieved.
    \item \textbf{No-Think}: Introduced in AdaptR1, this operator permits the model to bypass the explicit reasoning phase during a scheduled \textbf{No-Think} step.
\end{itemize}

\section{AdaptR1}

AdaptR1 extends GRPO to train an interleaved QA policy that can decide at each scheduled reasoning slot whether to generate an explicit rationale or emit the \textbf{No-Think} token \texttt{<think>no\_think</think>}. This step-wise decision is important for multi-hop QA: a trajectory may need explicit reasoning after some retrieval results, but not after every intermediate step. We train this behavior directly with RL rather than SFT, since constructing oracle trajectories that label exactly when reasoning should be skipped is ambiguous and dataset-dependent.

\subsection{Quality-Gated Efficiency Objective}

AdaptR1 is designed to optimize efficiency under an answer-quality constraint rather than as a pure length penalty. Let $R_{\text{ans}}(o)\in[0,1]$ denote the answer reward of a generated trajectory $o$, measured by F1, and let $R_{\text{nt}}(o)\in[0,1]$ denote a bounded efficiency reward derived from the number of No-Think actions. The intended objective is to reward efficiency only inside the feasible region of sufficiently accurate answers:
\begin{equation}
\max_{\pi_{\theta}} \mathbb{E}_{o\sim\pi_{\theta}}[R_{\text{nt}}(o)]
\quad \text{s.t.}\quad R_{\text{ans}}(o)\geq\tau.
\end{equation}
Operationally, we implement this constraint with a threshold mask and scale the efficiency bonus by the answer reward:
\begin{equation}
\begin{aligned}
r_{\text{AdaptR1}}(o)
&=R_{\text{ans}}(o)\bigl(1+\omega R_{\text{nt}}(o) \\
&\quad \cdot \mathbb{I}[R_{\text{ans}}(o)\geq\tau]\bigr).
\end{aligned}
\label{eq:constrainedreward}
\end{equation}
Eq.~\ref{eq:constrainedreward} is a compact view of the objective: the threshold mask prevents low-quality trajectories from receiving positive efficiency gradients, while the clipping and KL penalty in GRPO retain the trust-region-style stabilization of the base optimizer.

\subsection{AdaptR1 Reward}
As illustrated in Figure \ref{fig:AdaptR1-main}, our default implementation uses an absolute No-Think reward. Let $n_{\text{nt}}(o)$ be the number of No-Think actions in trajectory $o$. We first compute an uncapped efficiency bonus and then bound it by 1:
\begin{align}
    r_{\text{nt}}(o) &= n_{\text{nt}}(o) \times r_0, \\
    R_{\text{nt}}(o)&=\min(r_{\text{nt}}(o), 1),
\end{align}
where $r_0$ is the unit reward for skipping one reasoning round. The actual training reward is the implementation form of Eq.~\ref{eq:constrainedreward}:
\begin{equation}
\label{eq:AdaptR1reward}
\begin{aligned}
r_{\text{AdaptR1}}(o)
&=r_{\text{answer}}(o) + \mathbb{I}[r_{\text{answer}}(o)\geq\tau] \\
&\quad \cdot \omega r_{\text{answer}}(o) R_{\text{nt}}(o).
\end{aligned}
\end{equation}
Here, $\tau$ is the answer-quality gate and $\omega$ controls the strength of the efficiency bonus relative to the answer reward. The ceiling on $R_{\text{nt}}$ keeps the No-Think signal auxiliary, so the model is rewarded for concise trajectories only when answer quality remains acceptable. The hyperparameters $\tau$ and $\omega$ are evaluated in the ablation studies.

\begin{table*}[htbp]
\centering
\fontsize{7pt}{7.5pt}\selectfont
\setlength{\tabcolsep}{0.8mm}{
\resizebox{0.98\textwidth}{!}{\begin{tabular}{l cc  cc cc cc cc cc |ccc}
\toprule
\multirow{2.5}{*}{\textbf{Method}} & \multicolumn{2}{c}{\textbf{2Wiki.}} & \multicolumn{2}{c}{\textbf{HotpotQA}} & \multicolumn{2}{c}{\textbf{Musique}} & \multicolumn{2}{c}{\textbf{NQ}} & \multicolumn{2}{c}{\textbf{PopQA}} & \multicolumn{2}{c}{\textbf{TriviaQA}} & \multicolumn{3}{c}{\textbf{Avg.}} \\
\cmidrule(lr){2-3} \cmidrule(lr){4-5} \cmidrule(lr){6-7} \cmidrule(lr){8-9} \cmidrule(lr){10-11} \cmidrule(lr){12-13} \cmidrule(lr){14-16}
 & EM & F1 & EM & F1 & EM & F1 & EM & F1 & EM & F1 & EM & F1 & EM & F1 & R-S \\
\midrule
\multicolumn{16}{c}{\textbf{\textit{GPT-4o-mini}}} \\
\raisebox{-0.22\height}{\includegraphics[width=0.02\textwidth]{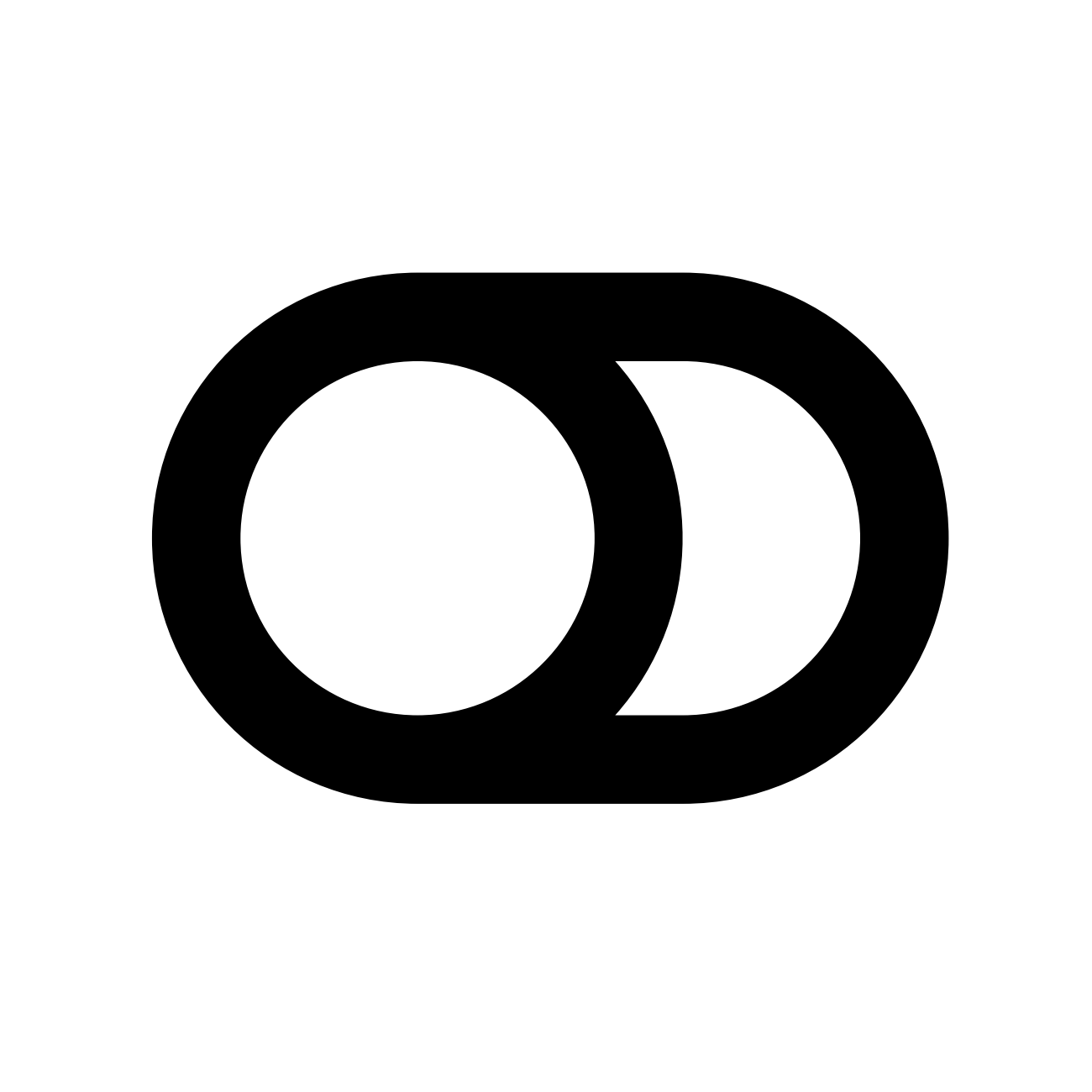}}\raisebox{-0.22\height}{\includegraphics[width=0.02\textwidth]{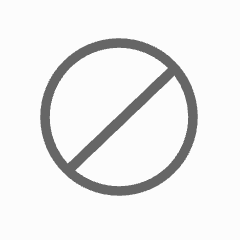}} NaiveGeneration & 4.69& 17.03& 18.75& 31.79& 3.13& 11.45& 2.34& 21.59& 10.36& 25.95& 28.91& 47.73
& 11.36 & 25.92 & - \\
\raisebox{-0.22\height}{\includegraphics[width=0.02\textwidth]{close.png}}\raisebox{-0.22\height}{\includegraphics[width=0.02\textwidth]{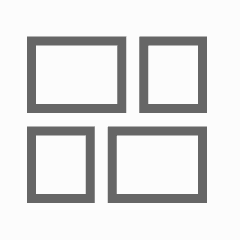}} StandardRAG & 7.03& 22.31& 35.16& 46.70& 9.38& 17.31& 7.03& 26.85& 18.75& 30.58& 31.25& 48.55
& 18.10 & 32.05 & 52.68 \\
\raisebox{-0.22\height}{\includegraphics[width=0.02\textwidth]{close.png}}\raisebox{-0.22\height}{\includegraphics[width=0.02\textwidth]{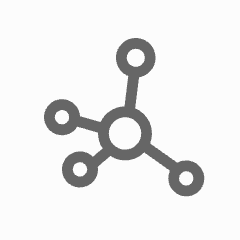}} GraphRAG & 3.91& 16.02& 19.53& 31.67& 7.03& 15.14& 3.91& 20.31& 8.59& 20.92& 32.03& 45.13
& 12.50 & 24.87 & 32.48  \\
\raisebox{-0.22\height}{\includegraphics[width=0.02\textwidth]{close.png}}\raisebox{-0.22\height}{\includegraphics[width=0.02\textwidth]{graph.png}} LightRAG & 3.13& 16.59& 18.75& 30.70& 3.91& 14.39& 2.34& 19.09& 5.47& 24.47& 25.00& 40.18
& 9.77 & 24.24 & 47.42\\
\raisebox{-0.22\height}{\includegraphics[width=0.02\textwidth]{close.png}}\raisebox{-0.22\height}{\includegraphics[width=0.02\textwidth]{graph.png}} PathRAG & 3.91& 12.42& 10.94& 23.12& 3.13& 11.49& 2.34& 20.01& 2.34& 15.65& 19.53& 37.44
& 7.03 & 20.02 & 46.71  \\
\raisebox{-0.22\height}{\includegraphics[width=0.02\textwidth]{close.png}}\raisebox{-0.22\height}{\includegraphics[width=0.02\textwidth]{graph.png}} HippoRAG2 & 7.03& 16.27 & 19.53 & 31.78 & 6.25& 12.37& 7.81& 24.56& 9.38& 21.10& 32.81& 48.86
& 13.80 & 25.82 & 36.41 \\
\raisebox{-0.22\height}{\includegraphics[width=0.02\textwidth]{close.png}}\raisebox{-0.22\height}{\includegraphics[width=0.02\textwidth]{graph.png}} HyperGraphRAG & 4.69 & 21.14 & 21.88 & 37.46 & 6.25 & 20.40 & 3.91 & 22.95 & 13.28 & 29.48 & 28.91 & 44.95 & 13.15 & 29.40 & 61.82 \\
\midrule
\multicolumn{16}{c}{\textbf{\textit{Qwen2.5-7B-Instruct}}} \\
\raisebox{-0.22\height}{\includegraphics[width=0.02\textwidth]{close.png}}\raisebox{-0.22\height}{\includegraphics[width=0.02\textwidth]{none.png}} NaiveGeneration & 3.12& 12.25& 6.25& 18.58& 0.00& 4.06& 1.56& 13.00& 0.78& 12.82& 7.03& 24.51
& 3.12 & 14.20 & -  \\
\raisebox{-0.22\height}{\includegraphics[width=0.02\textwidth]{close.png}}\raisebox{-0.22\height}{\includegraphics[width=0.02\textwidth]{chunk.png}} StandardRAG & 7.81& 12.75& 10.16& 21.10& 0.78& 4.53& 1.56& 15.97& 3.12& 13.10& 8.59& 24.90
& 5.34 & 15.39 & 52.67  \\
\raisebox{-0.22\height}{\includegraphics[width=0.02\textwidth]{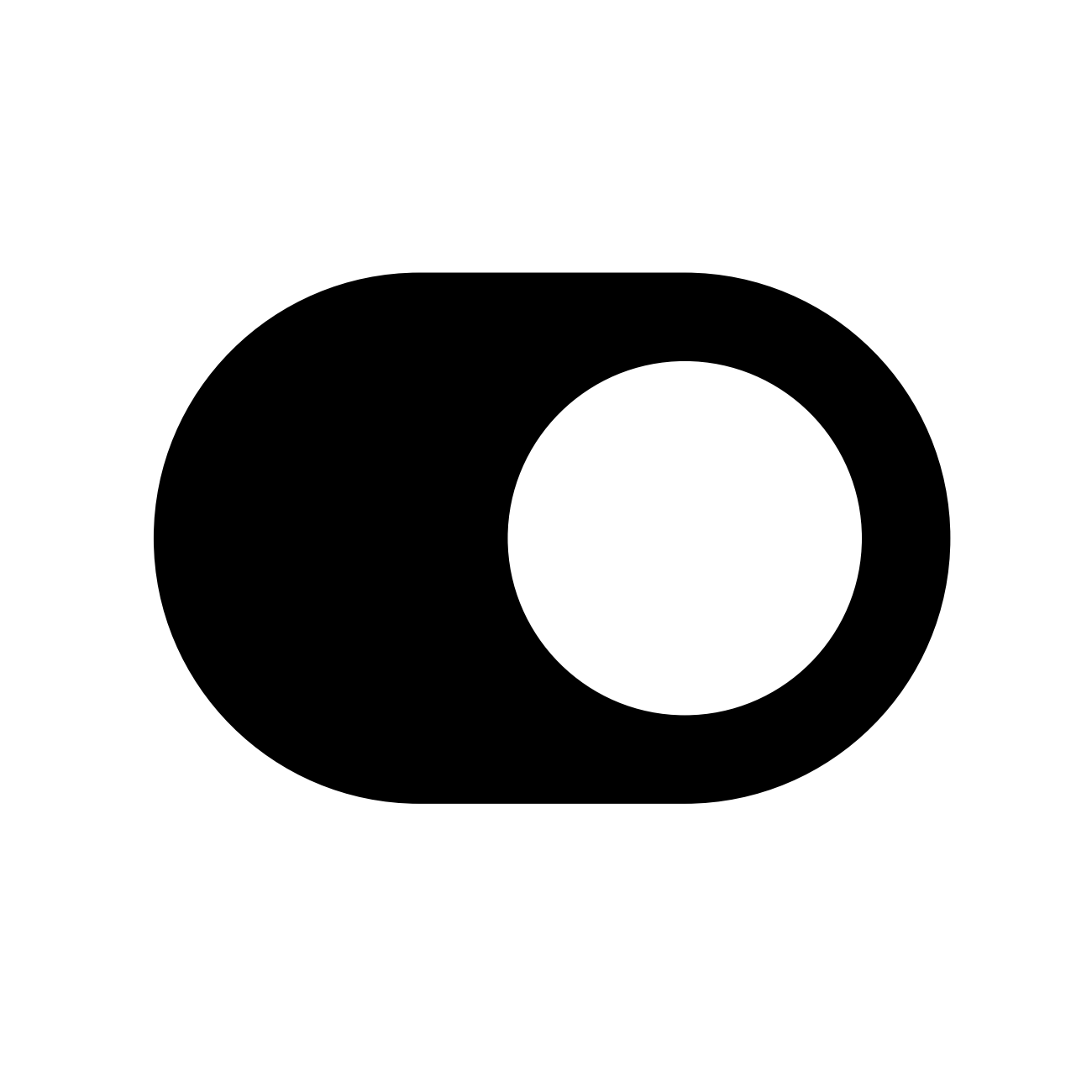}}\raisebox{-0.22\height}{\includegraphics[width=0.02\textwidth]{none.png}} SFT & 11.72& 20.28& 19.53& 27.59& 5.47& 10.02& 5.12& 19.02& 20.31& 27.93& 31.25& 39.21
& 15.57 & 24.01 & -  \\
 \raisebox{-0.22\height}{\includegraphics[width=0.02\textwidth]{open.png}}\raisebox{-0.22\height}{\includegraphics[width=0.02\textwidth]{none.png}} R1 & 25.00& 30.99& 31.25& 37.05& 7.03& 14.53& 16.41& 28.45& 26.56& 30.35& 49.22& 57.33
& 25.91 & 33.12 & -  \\
\raisebox{-0.22\height}{\includegraphics[width=0.02\textwidth]{open.png}}\raisebox{-0.22\height}{\includegraphics[width=0.02\textwidth]{chunk.png}} R1-Searcher & 27.34& 33.96& 39.84& 46.36& 10.16& 16.63& 32.03& 44.93& 41.41& 47.12& 56.25& 64.76
& 34.51 & 42.29 & 51.26  \\
\rowcolor{R1!15}\raisebox{-0.22\height}{\includegraphics[width=0.02\textwidth]{open.png}}\raisebox{-0.22\height}{\includegraphics[width=0.02\textwidth]{chunk.png}} Search-R1 & 35.15& 38.21& 43.77& 51.26& 17.18& 21.45& \textbf{38.34}& 43.79& 43.75& 47.03& 51.56& 61.03& 38.29 & 43.80 & 53.06 
\\
\rowcolor{R1!15}\raisebox{-0.22\height}{\includegraphics[width=0.02\textwidth]{open.png}}\raisebox{-0.22\height}{\includegraphics[width=0.02\textwidth]{chunk.png}} Search-AdaptR1 & 45.31& 51.55& 47.66& 53.68& 25.00& 34.31& 33.59& 47.13& 42.19& 46.80& 61.72& 70.63& 42.58& 50.68& \textbf{65.13} 
\\
\rowcolor{Graph-R1!15} \raisebox{-0.22\height}{\includegraphics[width=0.02\textwidth]{open.png}}\raisebox{-0.22\height}{\includegraphics[width=0.02\textwidth]{graph.png}} Graph-R1  & 58.59& 68.18& 55.47& 63.55& 37.50& 48.33& 35.16& 49.55& \textbf{50.78}& 54.01& 66.41& 72.02& 50.65& 59.27& 60.46 \\
\rowcolor{Graph-R1!15} \raisebox{-0.22\height}{\includegraphics[width=0.02\textwidth]{open.png}}\raisebox{-0.22\height}{\includegraphics[width=0.02\textwidth]{graph.png}} Graph-AdaptR1  & \textbf{61.72} & \textbf{69.20} & \textbf{57.81} & \textbf{64.39} & \textbf{39.84} & \textbf{53.42} & 35.94 & \textbf{49.62} & 49.22 & \textbf{55.03} & \textbf{64.06} & \textbf{72.77} & \textbf{51.43} & \textbf{60.74} & 61.88\\
% \midrule
% \multicolumn{16}{c}{\textbf{\textit{Qwen2.5-14B-Instruct}}} \\

% \rowcolor{Graph-R1!15} \raisebox{-0.22\height}{\includegraphics[width=0.02\textwidth]{open.png}}\raisebox{-0.22\height}{\includegraphics[width=0.02\textwidth]{graph.png}} Graph-R1  & 67.97& 75.46& 67.19& 72.52& 43.75& 57.54& 39.84& 53.81& 49.22& 53.33& 68.75& 76.43& 56.12& 64.85& 60.65
% \\
% \rowcolor{Graph-R1!15} \raisebox{-0.22\height}{\includegraphics[width=0.02\textwidth]{open.png}}\raisebox{-0.22\height}{\includegraphics[width=0.02\textwidth]{graph.png}} Graph-AdaptR1  & & & & & & & & & & & & & & & \\

\bottomrule
\end{tabular}}%
}
\caption{\label{T2} Main results under the controlled multi-hop QA setting with best in \textbf{bold}. \raisebox{-0.5mm}{\includegraphics[width=0.02\textwidth]{close.png}} means prompt engineering, \raisebox{-0.5mm}{\includegraphics[width=0.02\textwidth]{open.png}} means training, \raisebox{-0.5mm}{\includegraphics[width=0.02\textwidth]{none.png}} means no knowledge interaction, \raisebox{-0.5mm}{\includegraphics[width=0.02\textwidth]{chunk.png}} means chunk-based knowledge, and \raisebox{-0.5mm}{\includegraphics[width=0.02\textwidth]{graph.png}} means graph-based knowledge.}\label{tab:adaptr1}
\end{table*}

\subsection{Step-Wise Weighting and Reward Variants}

The main experiments use the absolute No-Think reward in Eq.~\ref{eq:AdaptR1reward}. To make this reward position-aware, we replace $R_{\text{nt}}(o)$ with a step-wise weighted efficiency reward. Let $n_{\text{nt}}^{(j)}(o)$ count No-Think actions at step $j$. Since the average trajectory length in our datasets is typically two to three rounds, we separate the first step from later steps:
\begin{equation}
\label{eq:weighted-nothink}
\begin{aligned}
s_{\lambda}(o)
&=\lambda n_{\text{nt}}^{(1)}(o)
 +(1-\lambda)\sum_{j>1} n_{\text{nt}}^{(j)}(o), \\
R_{\text{nt}}^{\lambda}(o)
&=\min(r_0 s_{\lambda}(o), 1).
\end{aligned}
\end{equation}
The coefficient $\lambda$ controls whether the reward pressure favors early or later No-Think actions. We set $\lambda=0.9$ in the main experiments and study its sensitivity in Section~\ref{sec:ablations}.

For reward-shape ablations, we additionally compare against a relative variant that normalizes the efficiency bonus by the total number of rounds:
\begin{align}
    p_{\text{nt}}(o) = \frac{n_{\text{nt}}(o)}{n_{\text{all}}(o)}.
\end{align}
This variant uses the same answer-quality gate as AdaptR1:
\begin{equation}
\label{eq:AdaptR1reward-relative}
\begin{aligned}
r_{\text{rel}}(o)
&=r_{\text{answer}}(o) + \mathbb{I}[r_{\text{answer}}(o)\geq\tau] \\
&\quad \cdot \omega r_{\text{answer}}(o) p_{\text{nt}}(o).
\end{aligned}
\end{equation}
This relative reward is not the default AdaptR1 objective; it is included to test whether ratio-based normalization encourages more stable exploration than the absolute bounded bonus.

\section{Experiments}

\subsection{Setups}
% \vspace{-3mm}

\textbf{Datasets and Metrics.} Following Graph-R1\citep{Graph-R1}, we conduct experiments on six common QA datasets\citep{FlashRAG}: 2Wikihop\citep{2WikiMultiHopQA}, HotpotQA\citep{HotpotQA}, Musique\citep{Musique}, NQ\citep{NQ}, PopQA\citep{PopQA}, and TriviaQA\citep{TriviaQA}. To keep comparisons controlled, all training-based methods use the same fixed split of 5,120 training and 128 testing instances per dataset. We use EM and F1 to evaluate answer quality, and R-S to evaluate retrieval performance.

\textbf{Baselines.} 
We consider both training-free and training-based baselines. 
The training-free methods include NaiveGeneration, StandardRAG~\citep{RAG}, GraphRAG~\citep{GraphRAG}, LightRAG~\citep{LightRAG}, PathRAG~\citep{PathRAG}, HippoRAG2~\citep{HippoRAG2}, and HyperGraphRAG~\citep{HyperGraphRAG}. 
The training-based baselines include SFT~\citep{SFT}, R1~\citep{GRPO}, R1-Searcher~\citep{R1-Searcher}, Search-R1~\citep{Search-R1}, and Graph-R1~\citep{Graph-R1}. 
For the key RL comparisons, Search-R1, Graph-R1, Search-AdaptR1, and Graph-AdaptR1 are initialized from the same Qwen2.5-7B-Instruct\citep{Qwen2.5} backbone and trained under the same data split and hyperparameter budget; AdaptR1 is not initialized from trained Search-R1 or Graph-R1 checkpoints. This isolates the effect of the adaptive reward from differences in data, initialization, and training overhead.

\textbf{Implementation Details.}
We instantiate AdaptR1 on two baselines, Search-R1 and Graph-R1, yielding Search-AdaptR1 and Graph-AdaptR1. 
Unless otherwise stated, we use the absolute No-Think reward with weight $\lambda=0.9$, threshold $\tau=0.6$ and coefficient $\omega=0.2$. Detailed implementations of baselines and our method are provided in Appendix~\ref{appendix:detailedimplementation}.

\textbf{Retriever.} The retriever follows the corresponding backbone method. Search-R1 uses E5\citep{wang2022text}, while Graph-R1 uses hypergraph-based retrieval with bge-large-en-v1.5\citep{BAAIembedding}.

\begin{table*}[htbp]
    \centering
    \small
    \resizebox{\textwidth}{!}{\begin{tabular}{lcccccc|c}
        \toprule
        Model
        & 2Wiki
        & HotpotQA
        & Musique
        & NQ
        & PopQA
        & TriviaQA
        & Avg. \\
        
        \midrule

        Search-R1
        & 155.00 & 100.74 & 117.23 & 95.56 & 105.78 & 92.63 & 111.16 \\

        Search-AdaptR1
        & 119.21\textcolor{green!50!black}{($\downarrow$23.09\%)} & 64.73\textcolor{green!50!black}{($\downarrow$35.75\%)} & 82.52\textcolor{green!50!black}{($\downarrow$29.61\%)} & 60.27\textcolor{green!50!black}{($\downarrow$36.93\%)} & 42.60\textcolor{green!50!black}{($\downarrow$59.73\%)} & 90.77\textcolor{green!50!black}{($\downarrow$2.01\%)} & 76.68\textcolor{green!50!black}{($\downarrow$31.02\%)} \\

        \midrule
        
        Graph-R1
        & 151.14 & 103.68 & 148.29 & 61.99 & 54.01 & 93.33 & 102.07 \\
        
        Graph-AdaptR1
        & 58.92\textcolor{green!50!black}{($\downarrow$61.02\%)} & 10.00\textcolor{green!50!black}{($\downarrow$90.35\%)} & 86.06\textcolor{green!50!black}{($\downarrow$41.97\%)} & 10.00\textcolor{green!50!black}{($\downarrow$83.87\%)} & 10.00\textcolor{green!50!black}{($\downarrow$81.48\%)} & 10.52\textcolor{green!50!black}{($\downarrow$88.73\%)} & 30.92\textcolor{green!50!black}{($\downarrow$69.71\%)} \\
        \bottomrule
    \end{tabular}}
    \caption{Comparisons of \textbf{Think Tokens} before and after applying AdaptR1 in Search-R1 and Graph-R1 settings.}
    \label{tab:ablation_tokens}
\end{table*}

\subsection{Main Experiments}
We observe that Graph-AdaptR1 yields results comparable to Graph-R1, achieving an average F1 improvement of 1.5. Search-AdaptR1 outperforms Search-R1 by a larger margin, with an average F1 increase of 6.9. This consistent improvement across datasets and retrieval pipelines suggests that AdaptR1 is not tied to a single dataset or retrieval design. The improved R-S scores further indicate that adaptive skipping can preserve, and in some cases improve, the retrieval behavior needed for accurate answers.

\textbf{Think Token Economy.}
To evaluate the think token economy of AdaptR1, we analyze the token consumption detailed in Table \ref{tab:ablation_tokens}. Under the Graph-R1 setting, Graph-AdaptR1 lowers average think tokens from 102.07 to 30.92, corresponding to a 69.71\% reduction, while maintaining or slightly improving average F1. The largest reduction appears on HotpotQA, where think tokens decrease from 103.68 to 10.00, a 90.35\% reduction. This drastic decrease in token usage does not come at the cost of accuracy, as shown in the main experiments. Instead, it highlights that Graph-R1 contains substantial redundant explicit reasoning steps. By eliminating these superfluous reasoning steps, AdaptR1 lowers computational cost and latency while maintaining or improving performance.

\subsection{Case Study}
To provide a granular understanding of the model's behavior, we present a comparison of generation trajectories with and without AdaptR1 in Section \ref{appendix:casestudy}. The case study illustrates that standard LLMs can exhibit ``over-thinking,'' generating exhaustive and sometimes circular reasoning chains even in multi-hop QA. In contrast, AdaptR1 bypasses redundant thinking while preserving the key reasoning link needed for the answer. We additionally analyze cases where skipping thinking hurts answer quality in Section~\ref{appendix:failureanalysis}. These qualitative results corroborate our quantitative efficiency results while making clear that \texttt{no\_think} is beneficial when used selectively rather than as an unconditional rule.

\section{Ablations and Analysis}
\label{sec:ablations}

In this section, we validate the main design choices of \textbf{AdaptR1} through a prompt-only control experiment, a temporal distribution analysis, and reward-design ablations. All analyses use Graph-R1 as the backbone.

\subsection{Necessity of RL Training}
\label{sec:rlnecessity}
To verify that adaptive skipping is learned through RL rather than triggered by the prompt alone, we compare the base model before training with the RL-trained Graph-AdaptR1 model under the same adaptive prompt. As shown in Table~\ref{tab:before-after-rl}, the base model rarely uses \texttt{no\_think} on complex datasets and performs poorly. RL training substantially improves both answer quality and adaptive skipping; for example, on Musique, the No-Think rate increases from 13.02\% to 50.67\%, while F1 improves from 8.41 to 53.42. This indicates that \texttt{no\_think} becomes useful only after the model learns how to integrate it into the reason-search-answer trajectory.

\begin{table*}[htbp]
\centering
\small
\resizebox{0.92\textwidth}{!}{\begin{tabular}{llcccccc}
\toprule
\textbf{Metric} & \textbf{Model Status} & \textbf{2Wiki} & \textbf{HotpotQA} & \textbf{Musique} & \textbf{NQ} & \textbf{PopQA} & \textbf{TriviaQA} \\
\midrule
\multirow{2}{*}{No-Think Rate (\%)} & Before Training & 13.05 & 11.22 & 13.02 & 23.28 & 31.11 & 20.41 \\
 & After Training & 61.74 & 100.00 & 50.67 & 100.00 & 100.00 & 99.61 \\
\midrule
\multirow{2}{*}{F1 Score} & Before Training & 13.51 & 19.15 & 8.41 & 13.46 & 17.18 & 26.67 \\
 & After Training & 69.20 & 64.39 & 53.42 & 49.62 & 55.03 & 72.77 \\
\bottomrule
\end{tabular}}
\caption{Comparison before and after RL training using the same adaptive prompt.}
\label{tab:before-after-rl}
\end{table*}

\begin{table*}[htbp]
    \centering
    \small
    \begin{tabular}{l|cccccc|cc}
    \toprule
             &\textbf{Step 1} & \textbf{Step 2} & \textbf{Step 3} & \textbf{Step 4} & \textbf{Step 5} & \textbf{Step 6} & \textbf{Avg. Ratio} &  \textbf{F1}\\
        \midrule
        \multicolumn{9}{l}{\textit{Impact of Step-wise Weight ($\lambda$)}}\\
        \midrule
        $\lambda=0.5$ & 1.0000 & 0.9688 & 0.0391 & 0.0444 & 0.0000 & 0.0000 & 0.5294 & 0.4923\\
        \midrule
        \multicolumn{9}{l}{\textit{Sensitivity Analysis}}\\
        $\lambda=0.1$ & 0.0000 & 0.0000 & 0.0000 & 0.0000 & 0.0000 & - & 0.0000 & 0.5135\\
        $\lambda=0.2$ & 0.0000 & 0.0000 & 0.0156 & 0.0000 & 0.0000 & 0.0000 & 0.0047 & 0.4925\\
        $\lambda=0.3$ & 1.0000 & 0.0000 & 0.9297 & 0.4955 & 0.5000 & 0.0000 & 0.6043 & 0.4923\\
        $\lambda=0.4$ & 0.0000 & 0.0709 & 0.9762 & 0.9043 & 0.6667 & 1.0000 & 0.4780 & 0.4814\\
        \midrule
        $\lambda=0.6$ & 1.0000 & 0.1797 & 0.0106 & 0.0000 & 0.0000 & - & 0.4053 & 0.4724\\
        $\lambda=0.7$ & 0.0000 & 0.0000 & 0.0000 & 0.0000 & - & - & 0.0000 & 0.5235\\
        $\lambda=0.8$ & 1.0000 & 1.0000 & 0.9453 & 0.9500 & 0.8696 & 1.0000 & \textbf{0.9705} & 0.4743\\
        $\lambda=0.9$ & 1.0000 & 0.0000 & 0.8359 & 0.1983 & 0.3333 & 0.0000 & 0.5067 & \textbf{0.5342}\\
        \bottomrule
    \end{tabular}
    \caption{Analysis of the No-Think ratio from step 1 to 6 and performance (F1) for Graph-R1 with and without AdaptR1 on Musique. The parameter $\lambda$ controls the step-wise penalty weight. We observe the temporal distribution of token savings across sequential reasoning steps.}
    \label{tab:ablation_step-wise-token-saving-musique}
\end{table*}
\subsection{Step-wise Adaptive Thinking and Temporal Distribution}
We first evaluate the temporal dynamics of the No-Think mechanism to understand how the model allocates its reasoning budget across different stages of the problem-solving trajectory. Table \ref{tab:ablation_step-wise-token-saving-musique} presents the average step-wise No-Think ratio with a balanced penalty factor ($\lambda=0.5$) for Musique. Other datasets are shown in \ref{appendix:nothinkratios}. Compared with saturated datasets such as HotpotQA, NQ, PopQA, and TriviaQA, Musique exposes more complex reasoning dynamics and is therefore the most diagnostic setting for studying where explicit thinking remains necessary.

\textbf{Balanced Strategy:} The results reveal a distinct, emergent behavior: the model predominantly learns to bypass extensive reasoning during the initial steps (Steps 1--2), reserving its computational budget for the final stages of the trajectory (Steps 3--4). This observation is somewhat counter-intuitive; one might expect the initial planning phase to require significant cognitive load. However, the data suggests that for multi-hop QA tasks, the model adopts a ``retrieve-then-reason'' strategy. The early steps likely involve schema activation or direct information retrieval that can be handled heuristically, whereas the final steps require synthesis and deduction to formulate the answer. We therefore examine whether larger early-step weights better match this behavior.

\textbf{Sensitivity to $\lambda$:} We further explore the impact of varying the step-wise reward weight $\lambda$ from 0.1 to 0.9.
\begin{itemize}
    \item \textit{Low $\lambda$ (0.1--0.4):} Assigning insufficient reward to early No-Think actions generally leads to lower F1, suggesting that the model may still over-reason on simple intermediate steps.
    \item \textit{High $\lambda$ (0.6--0.9):} Larger $\lambda$ often improves over low $\lambda$, and $\lambda=0.9$ gives the best F1 on Musique. The model consistently skips the first step but still uses explicit thinking later when retrieved evidence must be synthesized, supporting a retrieve-then-reason strategy rather than indiscriminate skipping.
\end{itemize}
These findings support the step-wise reward design, so we use $\lambda=0.9$ in the main experiments in Table \ref{tab:adaptr1}.

\subsection{Design of the Adaptive Reward Function}
The efficacy of Reinforcement Learning (RL) is heavily contingent on reward shaping. In this subsection, we investigate the specific design components of the AdaptR1 reward structure, including formulation (absolute vs. relative), constraints (ceilings), and hyperparameter sensitivity.

\begin{table}[htbp]
    \centering
    \resizebox{0.48\textwidth}{!}{\begin{tabular}{lcccccc|c}
    \toprule
             &2Wiki. & HotpotQA & Musique & NQ & PopQA & TriviaQA & \textbf{Avg.}\\
        \midrule
        AdaptR1 & 69.20 & 64.39 & 53.42 & 49.62 & 55.03 & 72.77 & \textbf{60.74} \\
        AdaptR1-rel & 64.32 & 66.19 & 45.12 & 48.94 & 53.46 & 71.44 & 58.25\\
        w/o Top Ceiling & 63.29 & 61.86 & 46.37 & 47.38 & 49.19 & 69.12 & 56.20 \\
        % \midrule
        \bottomrule
    \end{tabular}}
    \caption{F1 scores of Graph-AdaptR1 variants across varying datasets. We contrast the standard formulation against relative rewards and no ceiling rewards.}
    \label{tab:ablations-miscellaneous}
\end{table}

\paragraph{Absolute vs. Relative Reward Formulation}
We compare our absolute No-Think reward against a relative formulation (Table \ref{tab:ablations-miscellaneous}). A relative reward scales with the fraction of skipped steps, which appears smoother but is easier to exploit: the model can shorten the trajectory and skip a larger ratio of steps regardless of context. In contrast, the absolute formulation preserves the multi-turn reasoning structure and yields better average F1.

\paragraph{Impact of Reward Ceiling}
We further test the necessity of a ``top ceiling,'' a hard limit on the accumulation of efficiency rewards. As shown in the ``w/o top ceiling'' row of Table \ref{tab:ablations-miscellaneous}, removing this constraint degrades performance significantly. Without a ceiling, the efficiency reward can dominate the optimization landscape and encourage reward hacking. The ceiling keeps efficiency as a secondary objective that should not override answer accuracy.

\paragraph{Threshold Sensitivity ($\tau$)}
The threshold $\tau$ determines the confidence level required for the model to trigger a No-Think action. Table \ref{tab:ablations-threshold} illustrates the trade-off:
\begin{itemize}
    \item \textit{Low $\tau$ ($<0.6$):} A permissive threshold results in frequent, unjustified skipping of reasoning steps, harming performance (Avg F1 $\approx 55.3$).
    \item \textit{High $\tau$ ($>0.8$):} An overly strict threshold renders the No-Think reward too sparse. The model rarely attempts to skip, negating the efficiency benefits of AdaptR1.
\end{itemize}
Our results identify $\tau=0.6$ as the critical inflection point where the model reliably filters unnecessary reasoning without truncating valid cognitive processes.

\begin{table}[htbp]
    \centering
    \resizebox{0.48\textwidth}{!}{\begin{tabular}{lcccccc|c}
    \toprule
         $\tau$    & 2Wiki. & HotpotQA & Musique & NQ & PopQA & TriviaQA & \textbf{Avg.} \\
        \midrule
        0.2 & 65.79 & 62.29 & 48.77 & 49.12 & 52.45 & 72.14 & 58.43 \\
        0.4 & 65.22 & 65.38 & 47.69 & 49.90 & 55.89 & 70.14 & 59.04 \\
        0.6 & 69.20 & 64.39 & 53.42 & 49.62 & 55.03 & 72.77 & \textbf{60.74} \\
        0.8 & 64.79 & 64.53 & 44.14 & 50.58 & 56.15 & 70.20 & 58.40 \\
        1.0 & 67.37 & 67.13 & 46.40 & 48.99 & 55.29 & 70.52 & 59.28 \\
        \bottomrule
    \end{tabular}}
    \caption{Ablation study on the confidence threshold $\tau$ with the ratio fixed at 0.2. Performance peaks at $\tau=0.6$, suggesting a balance between aggressive skipping and conservative reasoning.}
    \label{tab:ablations-threshold}
\end{table}

\paragraph{Reward Coefficient ($\omega$)}
Finally, we analyze the magnitude of the efficiency reward relative to the correctness reward, controlled by coefficient $\omega$ (Table \ref{tab:ablations-ratio}). The data exhibits an inverted U-shaped curve. A small $\omega$ ($0.1$) provides a weak adaptive signal, while a large $\omega$ ($>0.4$) distracts optimization from the primary QA objective. We find that $\omega=0.2$ provides the best trade-off, suggesting that the efficiency signal should remain auxiliary to answer correctness.

\begin{table}[htbp]
    \centering
    \resizebox{0.48\textwidth}{!}{\begin{tabular}{lcccccc|c}
    \toprule
         $\omega$    &2Wiki. & HotpotQA & Musique & NQ & PopQA & TriviaQA & \textbf{Avg.}\\
        \midrule
        0.1 &  63.75 & 65.30 & 48.11 & 47.68 & 49.09 & 71.13 & 57.51 \\
        0.2 &  69.20 & 64.39 & 53.42 & 49.62 & 55.03 & 72.77 & \textbf{60.74} \\
        0.3 &  62.56 & 60.98 & 45.95 & 48.29 & 47.68 & 69.44 & 55.82 \\
        0.4 &  65.72 & 63.71 & 45.21 & 51.95 & 54.22 & 70.82 & 58.61 \\
        0.5 &  64.80 & 67.27 & 46.63 & 47.42 & 51.18 & 70.47 & 57.96 \\
        0.6 &  62.24 & 58.39 & 49.54 & 48.87 & 47.76 & 68.60 & 55.90 \\
        \bottomrule
    \end{tabular}}
\caption{Ablation study on the reward coefficient $\omega$. The optimal value $\omega=0.2$ indicates that efficiency rewards must be carefully scaled relative to answer accuracy rewards.}
    \label{tab:ablations-ratio}
\end{table}

Taken together, the ablations provide practical guardrails for avoiding reward hacking: $\tau$ should not be lower than 0.6, $\omega$ is most reliable around 0.1--0.2, and $\lambda$ is best treated as a step-wise pressure term rather than a universal instruction to skip. These ranges preserve the answer reward as the dominant objective while allowing the model to discover efficient trajectories.

\subsection{Training Dynamics}

\begin{figure}[htbp]
    \centering
    \includegraphics[width=\linewidth]{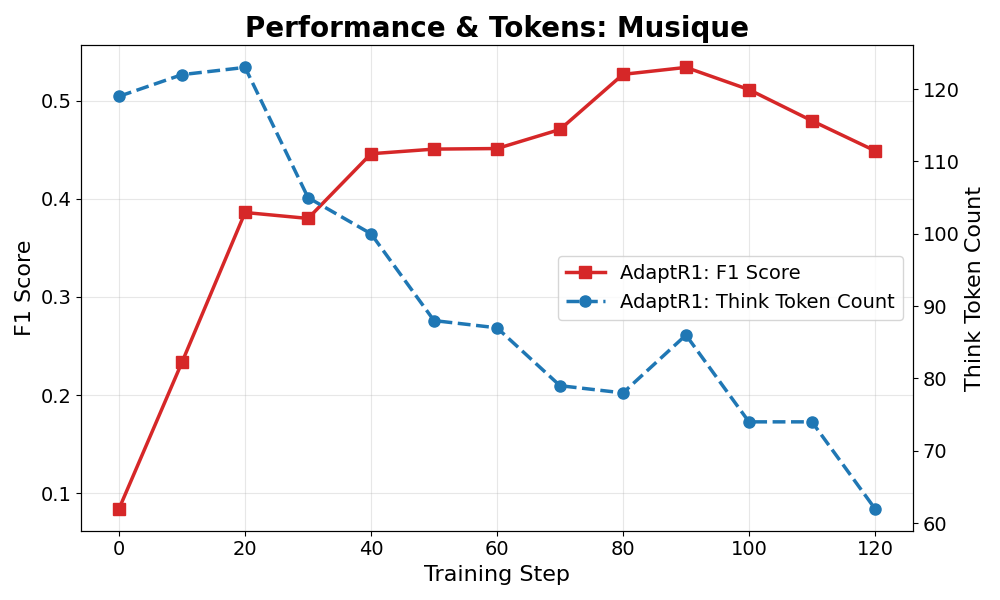}
    \caption{Training dynamics on Musique. The evolution of No-Think behavior indicates a phased learning process.}
    \label{fig:ablation-trainingdynamics}
\end{figure}

To better understand the learning progression of AdaptR1, we visualize No-Think behavior on the difficult Musique dataset in Figure \ref{fig:ablation-trainingdynamics}. The curve shows a phased rather than monotonic trade-off between accuracy and reasoning length. In early training (steps 0--20), $F_1$ rises rapidly to about 0.4 while think tokens remain high ($\sim 120$), suggesting that the model first learns to solve the task with ample reasoning. During steps 20--80, No-Think behavior emerges: $F_1$ peaks above 0.6 as think tokens decline toward $\sim 80$, showing that AdaptR1 learns to prune redundant thinking after acquiring task competence. Past step 90, however, tokens fall further to $\sim 60$ while $F_1$ drops to about 0.45, indicating that excessive pruning can skip necessary synthesis. Appendix \ref{appendix:trainingdynamics} shows the same trend on other datasets; most reduce think tokens early and then improve accuracy, while 2Wiki briefly recovers more thinking near step 80.

\section{Conclusion}

In this paper, we presented \textbf{AdaptR1}, an RL-based framework for mitigating ``over-thinking'' in multi-hop question answering. Unlike adaptive methods that rely on global routing decisions or SFT cold-start trajectories, AdaptR1 learns a fine-grained, step-wise policy that decides when to reason explicitly, when to query external knowledge, and when to skip redundant thinking at each intermediate stage. Empirically, AdaptR1 improves both Search-R1 and Graph-R1 settings: Search-AdaptR1 raises average F1 from 43.80 to 50.68, while Graph-AdaptR1 raises average F1 from 59.27 to 60.74. Under the Graph-R1 setting, it reduces average think tokens by 69.71\%, with the largest per-dataset reduction reaching 90.35\%, while maintaining or slightly improving answer performance. Our analyses further show that adaptive skipping is learned through RL, that overthinking is concentrated in the initial planning stages rather than the final synthesis stage, and that quality-gated rewards are important for avoiding reward hacking. These results suggest that RL-only adaptive interleaved thinking is a promising direction for efficient multi-hop reasoning.

\section*{Limitations}

The limitations of AdaptR1 primarily stem from its sensitivity to hyperparameters and potential training instability, as the method relies on balancing the confidence threshold ($\tau$), reward coefficient ($\omega$), and step-wise weights ($\lambda$). Our ablations identify useful ranges, but overly aggressive settings can still cause the model to over-prune essential reasoning steps and prioritize brevity over correctness in later training epochs. In addition, our current scope is multi-hop QA, where trajectories are typically short to medium length. We do not claim that the same reward design directly transfers to DeepResearch-style tasks that require much longer planning and sustained reasoning; extending AdaptR1 to that setting remains future work.


\bibliography{custom}

@misc{deng2025hipohybridpolicyoptimization,
      title={HiPO: Hybrid Policy Optimization for Dynamic Reasoning in LLMs}, 
      author={Ken Deng and Zizheng Zhan and Wen Xiang and Wenqiang Zhu and Weihao Li and Jingxuan Xu and Tianhao Peng and Xinping Lei and Kun Wu and Yifan Yao and Haoyang Huang and Huaixi Tang and Kepeng Lei and Zhiyi Lai and Songwei Yu and Zongxian Feng and Zuchen Gao and Weihao Xie and Chenchen Zhang and Yanan Wu and Yuanxing Zhang and Lecheng Huang and Yuqun Zhang and Jie Liu and Zhaoxiang Zhang and Haotian Zhang and Bin Chen and Jiaheng Liu},
      year={2025},
      eprint={2509.23967},
      archivePrefix={arXiv},
      primaryClass={cs.CL},
      url={https://arxiv.org/abs/2509.23967}, 
}

@misc{zhang2025adaptthinkreasoningmodelslearn,
      title={AdaptThink: Reasoning Models Can Learn When to Think}, 
      author={Jiajie Zhang and Nianyi Lin and Lei Hou and Ling Feng and Juanzi Li},
      year={2025},
      eprint={2505.13417},
      archivePrefix={arXiv},
      primaryClass={cs.CL},
      url={https://arxiv.org/abs/2505.13417}, 
}

@misc{tu2025learningthinkshapingadaptive,
      title={Learning When to Think: Shaping Adaptive Reasoning in R1-Style Models via Multi-Stage RL}, 
      author={Songjun Tu and Jiahao Lin and Qichao Zhang and Xiangyu Tian and Linjing Li and Xiangyuan Lan and Dongbin Zhao},
      year={2025},
      eprint={2505.10832},
      archivePrefix={arXiv},
      primaryClass={cs.CL},
      url={https://arxiv.org/abs/2505.10832}, 
}

@article{wu2025arm,
  title={ARM: Adaptive Reasoning Model},
  author={Wu, Siye and Xie, Jian and Zhang, Yikai and Chen, Aili and Zhang, Kai and Su, Yu and Xiao, Yanghua},
  journal={arXiv preprint arXiv:2505.20258},
  year={2025}
}

@article{wu2025unlocking,
  author       = {Han Wu and
                  Yuxuan Yao and
                  Shuqi Liu and
                  Zehua Liu and
                  Xiaojin Fu and
                  Xiongwei Han and
                  Xing Li and
                  Hui{-}Ling Zhen and
                  Tao Zhong and
                  Mingxuan Yuan},
  title        = {Unlocking Efficient Long-to-Short {LLM} Reasoning with Model Merging},
  journal      = {CoRR},
  volume       = {abs/2503.20641},
  year         = {2025},
  url          = {https://doi.org/10.48550/arXiv.2503.20641},
  doi          = {10.48550/ARXIV.2503.20641},
  eprinttype    = {arXiv},
  eprint       = {2503.20641},
  bibsource    = {dblp computer science bibliography, https://dblp.org}
}

@inproceedings{kang2025c3ot,
  title={C3ot: Generating shorter chain-of-thought without compromising effectiveness},
  author={Kang, Yu and Sun, Xianghui and Chen, Liangyu and Zou, Wei},
  booktitle={Proceedings of the AAAI Conference on Artificial Intelligence},
  volume={39},
  number={23},
  pages={24312--24320},
  year={2025}
}

@misc{ma2025cotvalvelengthcompressiblechainofthoughttuning,
      title={CoT-Valve: Length-Compressible Chain-of-Thought Tuning}, 
      author={Xinyin Ma and Guangnian Wan and Runpeng Yu and Gongfan Fang and Xinchao Wang},
      year={2025},
      eprint={2502.09601},
      archivePrefix={arXiv},
      primaryClass={cs.AI},
      url={https://arxiv.org/abs/2502.09601}, 
}

@misc{lou2025adacotparetooptimaladaptivechainofthought,
      title={AdaCoT: Pareto-Optimal Adaptive Chain-of-Thought Triggering via Reinforcement Learning}, 
      author={Chenwei Lou and Zewei Sun and Xinnian Liang and Meng Qu and Wei Shen and Wenqi Wang and Yuntao Li and Qingping Yang and Shuangzhi Wu},
      year={2025},
      eprint={2505.11896},
      archivePrefix={arXiv},
      primaryClass={cs.LG},
      url={https://arxiv.org/abs/2505.11896}, 
}

@misc{hou2025thinkprunepruninglongchainofthought,
      title={ThinkPrune: Pruning Long Chain-of-Thought of LLMs via Reinforcement Learning}, 
      author={Bairu Hou and Yang Zhang and Jiabao Ji and Yujian Liu and Kaizhi Qian and Jacob Andreas and Shiyu Chang},
      year={2025},
      eprint={2504.01296},
      archivePrefix={arXiv},
      primaryClass={cs.CL},
      url={https://arxiv.org/abs/2504.01296}, 
}

@article{aggarwal2025l1,
  title={L1: Controlling how long a reasoning model thinks with reinforcement learning},
  author={Aggarwal, Pranjal and Welleck, Sean},
  journal={arXiv preprint arXiv:2503.04697},
  year={2025}
}

@misc{chen2025a2fmadaptiveagentfoundation,
      title={A$^2$FM: An Adaptive Agent Foundation Model for Tool-Aware Hybrid Reasoning}, 
      author={Qianben Chen and Jingyi Cao and Jiayu Zhang and Tianrui Qin and Xiaowan Li and King Zhu and Dingfeng Shi and He Zhu and Minghao Liu and Xiaobo Liang and Xin Gui and Ge Zhang and Jian Yang and Yuchen Eleanor Jiang and Wangchunshu Zhou},
      year={2025},
      eprint={2510.12838},
      archivePrefix={arXiv},
      primaryClass={cs.CL},
      url={https://arxiv.org/abs/2510.12838}, 
}

@article{ma2025reasoning,
  title={Reasoning Models Can Be Effective Without Thinking},
  author={Ma, Wenjie and He, Jingxuan and Snell, Charlie and Griggs, Tyler and Min, Sewon and Zaharia, Matei},
  journal={arXiv preprint arXiv:2504.09858},
  year={2025}
}

@article{luo2025o1pruner,
  author       = {Haotian Luo and
                  Li Shen and
                  Haiying He and
                  Yibo Wang and
                  Shiwei Liu and
                  Wei Li and
                  Naiqiang Tan and
                  Xiaochun Cao and
                  Dacheng Tao},
  title        = {O1-Pruner: Length-Harmonizing Fine-Tuning for O1-Like Reasoning Pruning},
  journal      = {CoRR},
  volume       = {abs/2501.12570},
  year         = {2025},
  url          = {https://doi.org/10.48550/arXiv.2501.12570},
  doi          = {10.48550/ARXIV.2501.12570},
  eprinttype    = {arXiv},
  eprint       = {2501.12570},
  bibsource    = {dblp computer science bibliography, https://dblp.org}
}

@article{shen2025dast,
  author       = {Yi Shen and
                  Jian Zhang and
                  Jieyun Huang and
                  Shuming Shi and
                  Wenjing Zhang and
                  Jiangze Yan and
                  Ning Wang and
                  Kai Wang and
                  Shiguo Lian},
  title        = {{DAST:} Difficulty-Adaptive Slow-Thinking for Large Reasoning Models},
  journal      = {CoRR},
  volume       = {abs/2503.04472},
  year         = {2025},
  url          = {https://doi.org/10.48550/arXiv.2503.04472},
  doi          = {10.48550/ARXIV.2503.04472},
  eprinttype    = {arXiv},
  eprint       = {2503.04472},
  bibsource    = {dblp computer science bibliography, https://dblp.org}
}

@article{chen2025overthink,
  author       = {Xingyu Chen and
                  Jiahao Xu and
                  Tian Liang and
                  Zhiwei He and
                  Jianhui Pang and
                  Dian Yu and
                  Linfeng Song and
                  Qiuzhi Liu and
                  Mengfei Zhou and
                  Zhuosheng Zhang and
                  Rui Wang and
                  Zhaopeng Tu and
                  Haitao Mi and
                  Dong Yu},
  title        = {Do {NOT} Think That Much for 2+3=? On the Overthinking of o1-Like
                  LLMs},
  journal      = {CoRR},
  volume       = {abs/2412.21187},
  year         = {2024},
  url          = {https://doi.org/10.48550/arXiv.2412.21187},
  doi          = {10.48550/ARXIV.2412.21187},
  eprinttype    = {arXiv},
  eprint       = {2412.21187},
  bibsource    = {dblp computer science bibliography, https://dblp.org}
}

@article{kimiteam2025kimik15,
  author       = {Kimi Team and
                  Angang Du and
                  Bofei Gao and
                  Bowei Xing and
                  Changjiu Jiang and
                  Cheng Chen and
                  Cheng Li and
                  Chenjun Xiao and
                  Chenzhuang Du and
                  Chonghua Liao and
                  Chuning Tang and
                  Congcong Wang and
                  Dehao Zhang and
                  Enming Yuan and
                  Enzhe Lu and
                  Fengxiang Tang and
                  Flood Sung and
                  Guangda Wei and
                  Guokun Lai and
                  Haiqing Guo and
                  Han Zhu and
                  Hao Ding and
                  Hao Hu and
                  Hao Yang and
                  Hao Zhang and
                  Haotian Yao and
                  Haotian Zhao and
                  Haoyu Lu and
                  Haoze Li and
                  Haozhen Yu and
                  Hongcheng Gao and
                  Huabin Zheng and
                  Huan Yuan and
                  Jia Chen and
                  Jianhang Guo and
                  Jianlin Su and
                  Jianzhou Wang and
                  Jie Zhao and
                  Jin Zhang and
                  Jingyuan Liu and
                  Junjie Yan and
                  Junyan Wu and
                  Lidong Shi and
                  Ling Ye and
                  Longhui Yu and
                  Mengnan Dong and
                  Neo Zhang and
                  Ningchen Ma and
                  Qiwei Pan and
                  Qucheng Gong and
                  Shaowei Liu and
                  Shengling Ma and
                  Shupeng Wei and
                  Sihan Cao and
                  Siying Huang and
                  Tao Jiang and
                  Weihao Gao and
                  Weimin Xiong and
                  Weiran He and
                  Weixiao Huang and
                  Wenhao Wu and
                  Wenyang He and
                  Xianghui Wei and
                  Xianqing Jia and
                  Xingzhe Wu and
                  Xinran Xu and
                  Xinxing Zu and
                  Xinyu Zhou and
                  Xuehai Pan and
                  Y. Charles and
                  Yang Li and
                  Yangyang Hu and
                  Yangyang Liu and
                  Yanru Chen and
                  Yejie Wang and
                  Yibo Liu and
                  Yidao Qin and
                  Yifeng Liu and
                  Ying Yang and
                  Yiping Bao and
                  Yulun Du and
                  Yuxin Wu and
                  Yuzhi Wang and
                  Zaida Zhou and
                  Zhaoji Wang and
                  Zhaowei Li and
                  Zhen Zhu and
                  Zheng Zhang and
                  Zhexu Wang and
                  Zhilin Yang and
                  Zhiqi Huang and
                  Zihao Huang and
                  Ziyao Xu and
                  Zonghan Yang},
  title        = {Kimi k1.5: Scaling Reinforcement Learning with LLMs},
  journal      = {CoRR},
  volume       = {abs/2501.12599},
  year         = {2025},
  url          = {https://doi.org/10.48550/arXiv.2501.12599},
  doi          = {10.48550/ARXIV.2501.12599},
  eprinttype    = {arXiv},
  eprint       = {2501.12599},
  bibsource    = {dblp computer science bibliography, https://dblp.org}
}

@article{arora2025training,
  author       = {Daman Arora and
                  Andrea Zanette},
  title        = {Training Language Models to Reason Efficiently},
  journal      = {CoRR},
  volume       = {abs/2502.04463},
  year         = {2025},
  url          = {https://doi.org/10.48550/arXiv.2502.04463},
  doi          = {10.48550/ARXIV.2502.04463},
  eprinttype    = {arXiv},
  eprint       = {2502.04463},
  bibsource    = {dblp computer science bibliography, https://dblp.org}
}

@article{nayab2024concise,
  title={Concise thoughts: Impact of output length on llm reasoning and cost},
  author={Nayab, Sania and Rossolini, Giulio and Simoni, Marco and Saracino, Andrea and Buttazzo, Giorgio and Manes, Nicolamaria and Giacomelli, Fabrizio},
  journal={arXiv preprint arXiv:2407.19825},
  year={2024}
}

@article{sui2025stop,
  title={Stop overthinking: A survey on efficient reasoning for large language models},
  author={Sui, Yang and Chuang, Yu-Neng and Wang, Guanchu and Zhang, Jiamu and Zhang, Tianyi and Yuan, Jiayi and Liu, Hongyi and Wen, Andrew and Zhong, Shaochen and Zou, Na and others},
  journal={arXiv preprint arXiv:2503.16419},
  year={2025}
}

@article{kumar2025overthink,
  title={Overthink: Slowdown attacks on reasoning llms},
  author={Kumar, Abhinav and Roh, Jaechul and Naseh, Ali and Karpinska, Marzena and Iyyer, Mohit and Houmansadr, Amir and Bagdasarian, Eugene},
  journal={arXiv preprint arXiv:2502.02542},
  year={2025}
}

@article{Graph-R1,
  title={Graph-R1: Towards Agentic GraphRAG Framework via End-to-end Reinforcement Learning},
  author={Luo, Haoran and Chen, Guanting and Lin, Qika and Guo, Yikai and Xu, Fangzhi and Kuang, Zemin and Song, Meina and Wu, Xiaobao and Zhu, Yifan and Tuan, Luu Anh and others},
  journal={arXiv preprint arXiv:2507.21892},
  year={2025}
}

@misc{BAAIembedding,
  title={BGE M3-Embedding: Multi-Lingual, Multi-Functionality, Multi-Granularity Text Embeddings Through Self-Knowledge Distillation},
  author={Chen, Jianlv and Xiao, Shitao and Zhang, Peitian and Luo, Kun and Lian, Defu and Liu, Zheng},
  year={2023},
  eprint={2309.07597},
  archivePrefix={arXiv},
  primaryClass={cs.CL}
}

@inproceedings{SFT,
    title = "{L}lama{F}actory: Unified Efficient Fine-Tuning of 100+ Language Models",
    author = "Zheng, Yaowei  and
      Zhang, Richong  and
      Zhang, Junhao  and
      Ye, Yanhan  and
      Luo, Zheyan",
    editor = "Cao, Yixin  and
      Feng, Yang  and
      Xiong, Deyi",
    booktitle = "Proceedings of the 62nd Annual Meeting of the Association for Computational Linguistics (Volume 3: System Demonstrations)",
    month = aug,
    year = "2024",
    address = "Bangkok, Thailand",
    publisher = "Association for Computational Linguistics",
    url = "https://aclanthology.org/2024.acl-demos.38/",
    doi = "10.18653/v1/2024.acl-demos.38",
    pages = "400--410"
}

@misc{GPT-4o,
      title={GPT-4o System Card}, 
      author={OpenAI and : and Aaron Hurst and Adam Lerer and Adam P. Goucher and Adam Perelman and Aditya Ramesh and Aidan Clark and AJ Ostrow and Akila Welihinda and Alan Hayes and Alec Radford and Aleksander Mądry and Alex Baker-Whitcomb and Alex Beutel and Alex Borzunov and Alex Carney and Alex Chow and Alex Kirillov and Alex Nichol and Alex Paino and Alex Renzin and Alex Tachard Passos and Alexander Kirillov and Alexi Christakis and Alexis Conneau and Ali Kamali and Allan Jabri and Allison Moyer and Allison Tam and Amadou Crookes and Amin Tootoochian and Amin Tootoonchian and Ananya Kumar and Andrea Vallone and Andrej Karpathy and Andrew Braunstein and Andrew Cann and Andrew Codispoti and others},
      year={2024},
      eprint={2410.21276},
      archivePrefix={arXiv},
      primaryClass={cs.CL},
      url={https://arxiv.org/abs/2410.21276}, 
}

@inproceedings{RAG,
 author = {Lewis, Patrick and Perez, Ethan and Piktus, Aleksandra and Petroni, Fabio and Karpukhin, Vladimir and Goyal, Naman and K\"{u}ttler, Heinrich and Lewis, Mike and Yih, Wen-tau and Rockt\"{a}schel, Tim and Riedel, Sebastian and Kiela, Douwe},
 booktitle = {Advances in Neural Information Processing Systems},
 editor = {H. Larochelle and M. Ranzato and R. Hadsell and M.F. Balcan and H. Lin},
 pages = {9459--9474},
 publisher = {Curran Associates, Inc.},
 title = {Retrieval-Augmented Generation for Knowledge-Intensive NLP Tasks},
 url = {https://proceedings.neurips.cc/paper_files/paper/2020/file/6b493230205f780e1bc26945df7481e5-Paper.pdf},
 volume = {33},
 year = {2020}
}

@misc{GraphRAG,
      title={From Local to Global: A Graph RAG Approach to Query-Focused Summarization}, 
      author={Darren Edge and Ha Trinh and Newman Cheng and Joshua Bradley and Alex Chao and Apurva Mody and Steven Truitt and Dasha Metropolitansky and Robert Osazuwa Ness and Jonathan Larson},
      year={2025},
      eprint={2404.16130},
      archivePrefix={arXiv},
      primaryClass={cs.CL},
      url={https://arxiv.org/abs/2404.16130}, 
}

@misc{LightRAG,
      title={LightRAG: Simple and Fast Retrieval-Augmented Generation}, 
      author={Zirui Guo and Lianghao Xia and Yanhua Yu and Tu Ao and Chao Huang},
      year={2025},
      eprint={2410.05779},
      archivePrefix={arXiv},
      primaryClass={cs.IR},
      url={https://arxiv.org/abs/2410.05779}, 
}

@misc{HyperGraphRAG,
      title={HyperGraphRAG: Retrieval-Augmented Generation via Hypergraph-Structured Knowledge Representation}, 
      author={Haoran Luo and Haihong E and Guanting Chen and Yandan Zheng and Xiaobao Wu and Yikai Guo and Qika Lin and Yu Feng and Zemin Kuang and Meina Song and Yifan Zhu and Luu Anh Tuan},
      year={2025},
      eprint={2503.21322},
      archivePrefix={arXiv},
      primaryClass={cs.AI},
      url={https://arxiv.org/abs/2503.21322}, 
}

@misc{PathRAG,
      title={PathRAG: Pruning Graph-based Retrieval Augmented Generation with Relational Paths}, 
      author={Boyu Chen and Zirui Guo and Zidan Yang and Yuluo Chen and Junze Chen and Zhenghao Liu and Chuan Shi and Cheng Yang},
      year={2025},
      eprint={2502.14902},
      archivePrefix={arXiv},
      primaryClass={cs.CL},
      url={https://arxiv.org/abs/2502.14902}, 
}

@misc{HippoRAG2,
      title={From RAG to Memory: Non-Parametric Continual Learning for Large Language Models}, 
      author={Bernal Jiménez Gutiérrez and Yiheng Shu and Weijian Qi and Sizhe Zhou and Yu Su},
      year={2025},
      eprint={2502.14802},
      archivePrefix={arXiv},
      primaryClass={cs.CL},
      url={https://arxiv.org/abs/2502.14802}, 
}

@misc{GPT-o1,
      title={OpenAI o1 System Card}, 
      author={OpenAI and : and Aaron Jaech and Adam Kalai and Adam Lerer and Adam Richardson and Ahmed El-Kishky and Aiden Low and Alec Helyar and Aleksander Madry and Alex Beutel and Alex Carney and Alex Iftimie and Alex Karpenko and Alex Tachard Passos and Alexander Neitz and Alexander Prokofiev and Alexander Wei and Allison Tam and Ally Bennett and Ananya Kumar and Andre Saraiva and Andrea Vallone and Andrew Duberstein and Andrew Kondrich and Andrey Mishchenko and Andy Applebaum and Angela Jiang and Ashvin Nair and Barret Zoph and Behrooz Ghorbani and Ben Rossen and Benjamin Sokolowsky and Boaz Barak and Bob McGrew and Borys Minaiev and Botao Hao and others},
      year={2024},
      eprint={2412.16720},
      archivePrefix={arXiv},
      primaryClass={cs.AI},
      url={https://arxiv.org/abs/2412.16720}, 
}

@misc{GRPO,
      title={DeepSeekMath: Pushing the Limits of Mathematical Reasoning in Open Language Models}, 
      author={Zhihong Shao and Peiyi Wang and Qihao Zhu and Runxin Xu and Junxiao Song and Xiao Bi and Haowei Zhang and Mingchuan Zhang and Y. K. Li and Y. Wu and Daya Guo},
      year={2024},
      eprint={2402.03300},
      archivePrefix={arXiv},
      primaryClass={cs.CL},
      url={https://arxiv.org/abs/2402.03300}, 
}

@misc{Search-R1,
      title={Search-R1: Training LLMs to Reason and Leverage Search Engines with Reinforcement Learning}, 
      author={Bowen Jin and Hansi Zeng and Zhenrui Yue and Jinsung Yoon and Sercan Arik and Dong Wang and Hamed Zamani and Jiawei Han},
      year={2025},
      eprint={2503.09516},
      archivePrefix={arXiv},
      primaryClass={cs.CL},
      url={https://arxiv.org/abs/2503.09516}, 
}

@misc{R1-Searcher,
      title={R1-Searcher: Incentivizing the Search Capability in LLMs via Reinforcement Learning}, 
      author={Huatong Song and Jinhao Jiang and Yingqian Min and Jie Chen and Zhipeng Chen and Wayne Xin Zhao and Lei Fang and Ji-Rong Wen},
      year={2025},
      eprint={2503.05592},
      archivePrefix={arXiv},
      primaryClass={cs.AI},
      url={https://arxiv.org/abs/2503.05592}, 
}

@inproceedings{FlashRAG,
author = {Jin, Jiajie and Zhu, Yutao and Dou, Zhicheng and Dong, Guanting and Yang, Xinyu and Zhang, Chenghao and Zhao, Tong and Yang, Zhao and Wen, Ji-Rong},
title = {FlashRAG: A Modular Toolkit for Efficient Retrieval-Augmented Generation Research},
year = {2025},
isbn = {9798400713316},
publisher = {Association for Computing Machinery},
address = {New York, NY, USA},
url = {https://doi.org/10.1145/3701716.3715313},
doi = {10.1145/3701716.3715313},
booktitle = {Companion Proceedings of the ACM on Web Conference 2025},
pages = {737–740},
numpages = {4},
location = {Sydney NSW, Australia},
series = {WWW '25}
}

@inproceedings{2WikiMultiHopQA,
    title = "Constructing A Multi-hop {QA} Dataset for Comprehensive Evaluation of Reasoning Steps",
    author = "Ho, Xanh  and
      Duong Nguyen, Anh-Khoa  and
      Sugawara, Saku  and
      Aizawa, Akiko",
    editor = "Scott, Donia  and
      Bel, Nuria  and
      Zong, Chengqing",
    booktitle = "Proceedings of the 28th International Conference on Computational Linguistics",
    month = dec,
    year = "2020",
    address = "Barcelona, Spain (Online)",
    publisher = "International Committee on Computational Linguistics",
    url = "https://aclanthology.org/2020.coling-main.580/",
    doi = "10.18653/v1/2020.coling-main.580",
    pages = "6609--6625"
}

@inproceedings{HotpotQA,
    title = "{H}otpot{QA}: A Dataset for Diverse, Explainable Multi-hop Question Answering",
    author = "Yang, Zhilin  and
      Qi, Peng  and
      Zhang, Saizheng  and
      Bengio, Yoshua  and
      Cohen, William  and
      Salakhutdinov, Ruslan  and
      Manning, Christopher D.",
    editor = "Riloff, Ellen  and
      Chiang, David  and
      Hockenmaier, Julia  and
      Tsujii, Jun{'}ichi",
    booktitle = "Proceedings of the 2018 Conference on Empirical Methods in Natural Language Processing",
    month = oct # "-" # nov,
    year = "2018",
    address = "Brussels, Belgium",
    publisher = "Association for Computational Linguistics",
    url = "https://aclanthology.org/D18-1259/",
    doi = "10.18653/v1/D18-1259",
    pages = "2369--2380"
}

@article{Musique,
    title = "{M}u{S}i{Q}ue: Multihop Questions via Single-hop Question Composition",
    author = "Trivedi, Harsh  and
      Balasubramanian, Niranjan  and
      Khot, Tushar  and
      Sabharwal, Ashish",
    editor = "Roark, Brian  and
      Nenkova, Ani",
    journal = "Transactions of the Association for Computational Linguistics",
    volume = "10",
    year = "2022",
    address = "Cambridge, MA",
    publisher = "MIT Press",
    url = "https://aclanthology.org/2022.tacl-1.31/",
    doi = "10.1162/tacl_a_00475",
    pages = "539--554"
}

@article{NQ,
    title = "Natural Questions: A Benchmark for Question Answering Research",
    author = "Kwiatkowski, Tom  and
      Palomaki, Jennimaria  and
      Redfield, Olivia  and
      Collins, Michael  and
      Parikh, Ankur  and
      Alberti, Chris  and
      Epstein, Danielle  and
      Polosukhin, Illia  and
      Devlin, Jacob  and
      Lee, Kenton  and
      Toutanova, Kristina  and
      Jones, Llion  and
      Kelcey, Matthew  and
      Chang, Ming-Wei  and
      Dai, Andrew M.  and
      Uszkoreit, Jakob  and
      Le, Quoc  and
      Petrov, Slav",
    editor = "Lee, Lillian  and
      Johnson, Mark  and
      Roark, Brian  and
      Nenkova, Ani",
    journal = "Transactions of the Association for Computational Linguistics",
    volume = "7",
    year = "2019",
    address = "Cambridge, MA",
    publisher = "MIT Press",
    url = "https://aclanthology.org/Q19-1026/",
    doi = "10.1162/tacl_a_00276",
    pages = "452--466"
}

@inproceedings{PopQA,
    title = "When Not to Trust Language Models: Investigating Effectiveness of Parametric and Non-Parametric Memories",
    author = "Mallen, Alex  and
      Asai, Akari  and
      Zhong, Victor  and
      Das, Rajarshi  and
      Khashabi, Daniel  and
      Hajishirzi, Hannaneh",
    editor = "Rogers, Anna  and
      Boyd-Graber, Jordan  and
      Okazaki, Naoaki",
    booktitle = "Proceedings of the 61st Annual Meeting of the Association for Computational Linguistics (Volume 1: Long Papers)",
    month = jul,
    year = "2023",
    address = "Toronto, Canada",
    publisher = "Association for Computational Linguistics",
    url = "https://aclanthology.org/2023.acl-long.546/",
    doi = "10.18653/v1/2023.acl-long.546",
    pages = "9802--9822"
}

@inproceedings{TriviaQA,
    title = "{T}rivia{QA}: A Large Scale Distantly Supervised Challenge Dataset for Reading Comprehension",
    author = "Joshi, Mandar  and
      Choi, Eunsol  and
      Weld, Daniel  and
      Zettlemoyer, Luke",
    editor = "Barzilay, Regina  and
      Kan, Min-Yen",
    booktitle = "Proceedings of the 55th Annual Meeting of the Association for Computational Linguistics (Volume 1: Long Papers)",
    month = jul,
    year = "2017",
    address = "Vancouver, Canada",
    publisher = "Association for Computational Linguistics",
    url = "https://aclanthology.org/P17-1147/",
    doi = "10.18653/v1/P17-1147",
    pages = "1601--1611"
}

@misc{Qwen2.5,
      title={Qwen2.5 Technical Report}, 
      author={Qwen and : and An Yang and Baosong Yang and Beichen Zhang and Binyuan Hui and Bo Zheng and Bowen Yu and Chengyuan Li and Dayiheng Liu and Fei Huang and Haoran Wei and Huan Lin and Jian Yang and Jianhong Tu and Jianwei Zhang and Jianxin Yang and Jiaxi Yang and Jingren Zhou and Junyang Lin and Kai Dang and Keming Lu and Keqin Bao and Kexin Yang and Le Yu and Mei Li and Mingfeng Xue and Pei Zhang and Qin Zhu and Rui Men and Runji Lin and Tianhao Li and Tianyi Tang and Tingyu Xia and Xingzhang Ren and Xuancheng Ren and Yang Fan and Yang Su and Yichang Zhang and Yu Wan and Yuqiong Liu and Zeyu Cui and Zhenru Zhang and Zihan Qiu},
      year={2025},
      eprint={2412.15115},
      archivePrefix={arXiv},
      primaryClass={cs.CL},
      url={https://arxiv.org/abs/2412.15115}, 
}

@article{wei2022chain,
  title={Chain-of-thought prompting elicits reasoning in large language models},
  author={Wei, Jason and Wang, Xuezhi and Schuurmans, Dale and Bosma, Maarten and Xia, Fei and Chi, Ed and Le, Quoc V and Zhou, Denny and others},
  journal={Advances in neural information processing systems},
  volume={35},
  pages={24824--24837},
  year={2022}
}

@article{trivedi2022interleaving,
  title={Interleaving retrieval with chain-of-thought reasoning for knowledge-intensive multi-step questions},
  author={Trivedi, Harsh and Balasubramanian, Niranjan and Khot, Tushar and Sabharwal, Ashish},
  journal={arXiv preprint arXiv:2212.10509},
  year={2022}
}

@article{wang2022text,
  title={Text embeddings by weakly-supervised contrastive pre-training},
  author={Wang, Liang and Yang, Nan and Huang, Xiaolong and Jiao, Binxing and Yang, Linjun and Jiang, Daxin and Majumder, Rangan and Wei, Furu},
  journal={arXiv preprint arXiv:2212.03533},
  year={2022}
}

@inproceedings{jiang2023active,
  title={Active retrieval augmented generation},
  author={Jiang, Zhengbao and Xu, Frank F and Gao, Luyu and Sun, Zhiqing and Liu, Qian and Dwivedi-Yu, Jane and Yang, Yiming and Callan, Jamie and Neubig, Graham},
  booktitle={Proceedings of the 2023 Conference on Empirical Methods in Natural Language Processing},
  pages={7969--7992},
  year={2023}
}

@inproceedings{yao2023react,
  title={React: Synergizing reasoning and acting in language models},
  author={Yao, Shunyu and Zhao, Jeffrey and Yu, Dian and Du, Nan and Shafran, Izhak and Narasimhan, Karthik and Cao, Yuan},
  booktitle={International Conference on Learning Representations (ICLR)},
  year={2023}
}

@article{asai2024self,
  title={Self-rag: Learning to retrieve, generate, and critique through self-reflection},
  author={Asai, Akari and Wu, Zeqiu and Wang, Yizhong and Sil, Avirup and Hajishirzi, Hannaneh},
  year={2024},
  publisher={ICLR}
}

@article{DeepRAG,
  title={DeepRAG: Thinking to Retrieve Step by Step for Large Language Models},
  author={Guan, Xinyan and Zeng, Jiali and Meng, Fandong and Xin, Chunlei and Lu, Yaojie and Lin, Hongyu and Han, Xianpei and Sun, Le and Zhou, Jie},
  journal={arXiv preprint arXiv:2502.01142},
  year={2025}
}

@article{r3-rag,
  title={R3-RAG: Learning Step-by-Step Reasoning and Retrieval for LLMs via Reinforcement Learning},
  author={Li, Yuan and Luo, Qi and Li, Xiaonan and Li, Bufan and Cheng, Qinyuan and Wang, Bo and Zheng, Yining and Wang, Yuxin and Yin, Zhangyue and Qiu, Xipeng},
  journal={arXiv preprint arXiv:2505.23794},
  year={2025}
}

@misc{shao2023enhancing,
      title={Enhancing Retrieval-Augmented Large Language Models with Iterative Retrieval-Generation Synergy}, 
      author={Zhihong Shao and Yeyun Gong and Yelong Shen and Minlie Huang and Nan Duan and Weizhu Chen},
      year={2023},
      eprint={2305.15294},
      archivePrefix={arXiv},
      primaryClass={cs.CL},
      url={https://arxiv.org/abs/2305.15294}, 
}

@article{nakano2021webgpt,
  title={Webgpt: Browser-assisted question-answering with human feedback},
  author={Nakano, Reiichiro and Hilton, Jacob and Balaji, Suchir and Wu, Jeff and Ouyang, Long and Kim, Christina and Hesse, Christopher and Jain, Shantanu and Kosaraju, Vineet and Saunders, William and others},
  journal={arXiv preprint arXiv:2112.09332},
  year={2021}
}

@misc{press2023measuringnarrowingcompositionalitygap,
      title={Measuring and Narrowing the Compositionality Gap in Language Models}, 
      author={Ofir Press and Muru Zhang and Sewon Min and Ludwig Schmidt and Noah A. Smith and Mike Lewis},
      year={2023},
      eprint={2210.03350},
      archivePrefix={arXiv},
      primaryClass={cs.CL},
      url={https://arxiv.org/abs/2210.03350}, 
}

@inproceedings{zheng-etal-2025-deepresearcher,
    title = "{D}eep{R}esearcher: Scaling Deep Research via Reinforcement Learning in Real-world Environments",
    author = "Zheng, Yuxiang  and
      Fu, Dayuan  and
      Hu, Xiangkun  and
      Cai, Xiaojie  and
      Ye, Lyumanshan  and
      Lu, Pengrui  and
      Liu, Pengfei",
    editor = "Christodoulopoulos, Christos  and
      Chakraborty, Tanmoy  and
      Rose, Carolyn  and
      Peng, Violet",
    booktitle = "Proceedings of the 2025 Conference on Empirical Methods in Natural Language Processing",
    month = nov,
    year = "2025",
    address = "Suzhou, China",
    publisher = "Association for Computational Linguistics",
    url = "https://aclanthology.org/2025.emnlp-main.22/",
    doi = "10.18653/v1/2025.emnlp-main.22",
    pages = "414--431",
    ISBN = "979-8-89176-332-6"
}

@inproceedings{verl, series={EuroSys ’25},
   title={HybridFlow: A Flexible and Efficient RLHF Framework},
   url={http://dx.doi.org/10.1145/3689031.3696075},
   DOI={10.1145/3689031.3696075},
   booktitle={Proceedings of the Twentieth European Conference on Computer Systems},
   publisher={ACM},
   author={Sheng, Guangming and Zhang, Chi and Ye, Zilingfeng and Wu, Xibin and Zhang, Wang and Zhang, Ru and Peng, Yanghua and Lin, Haibin and Wu, Chuan},
   year={2025},
   month=mar, pages={1279–1297},
   collection={EuroSys ’25} }
\newpage
\appendix

\section{Prompt Template}
\label{appendix:prompt}

Table~\ref{tab:promptofAdaptR1} shows the Graph-AdaptR1 prompt. The template preserves the original reason-search-answer format while adding one adaptive instruction: when explicit reasoning is unnecessary, the model may emit \texttt{<think>no\_think</think>} before either a query or final answer. Search-AdaptR1 uses the same adaptive instruction but keeps Search-R1's native tool tags.

\begin{table*}[htbp]
\centering
\small
\begin{tabular}{p{0.98\textwidth}}
\toprule
Answer the given question. You can query from the knowledge base provided to you to answer the question. You can query knowledge as many times as you want.
You can conduct reasoning inside \textcolor{blue}{\texttt{\textbf{<think>}}}...\textcolor{blue}{\texttt{\textbf{</think>}}} when needed. \textcolor{red}{If reasoning is not necessary, output \textcolor{blue}{\texttt{\textbf{<think>}}}no\_think\textcolor{blue}{\texttt{\textbf{</think>}}} to skip reasoning.}
If you need to query knowledge, set \texttt{\{"query": <statement-to-search>\}} between \textcolor{orange}{\texttt{\textbf{<query>}}}...\textcolor{orange}{\texttt{\textbf{</query>}}} after the \textcolor{blue}{\texttt{\textbf{<think>}}}...\textcolor{blue}{\texttt{\textbf{</think>}}} tags.
When you have the final answer, output it inside \textcolor{purple}{\texttt{\textbf{<answer>}}}...\textcolor{purple}{\texttt{\textbf{</answer>}}} after the \textcolor{blue}{\texttt{\textbf{<think>}}}...\textcolor{blue}{\texttt{\textbf{</think>}}} tags. Please keep the answer short and clear.
\textbf{Formats:} tool call with reasoning: \textcolor{blue}{\texttt{\textbf{<think>}}}...\textcolor{blue}{\texttt{\textbf{</think>}}} \textcolor{orange}{\texttt{\textbf{<query>}}}\texttt{\{"query": <statement-to-search>\}}\textcolor{orange}{\texttt{\textbf{</query>}}}; tool call without reasoning: \textcolor{blue}{\texttt{\textbf{<think>}}}no\_think\textcolor{blue}{\texttt{\textbf{</think>}}} \textcolor{orange}{\texttt{\textbf{<query>}}}\texttt{\{"query": <statement-to-search>\}}\textcolor{orange}{\texttt{\textbf{</query>}}}; answer with reasoning: \textcolor{blue}{\texttt{\textbf{<think>}}}...\textcolor{blue}{\texttt{\textbf{</think>}}} \textcolor{purple}{\texttt{\textbf{<answer>}}}...\textcolor{purple}{\texttt{\textbf{</answer>}}}; answer without reasoning: \textcolor{blue}{\texttt{\textbf{<think>}}}no\_think\textcolor{blue}{\texttt{\textbf{</think>}}} \textcolor{purple}{\texttt{\textbf{<answer>}}}...\textcolor{purple}{\texttt{\textbf{</answer>}}}. Question: \textbf{question}. Assistant: \\
\bottomrule
\end{tabular}
\caption{Prompt template for Graph-AdaptR1. The red instruction introduces the adaptive No-Think action. Search-AdaptR1 uses the same adaptive instruction while preserving Search-R1's native \texttt{<search>}/\texttt{<information>} interface.}
\label{tab:promptofAdaptR1}
\end{table*}

\section{Ethics Statement}
This work utilizes publicly available datasets (2WikiMultiHopQA, HotpotQA, Musique, NQ, PopQA, and TriviaQA) that are widely used in the research community. We have adhered to the licenses and terms of use associated with these datasets. To the best of our knowledge, these datasets do not contain personally identifiable information (PII) or offensive content that would pose a risk to individuals. This study does not involve human subjects or human annotation, as all evaluations were conducted using automatic metrics.

A primary contribution of this work is the reduction of computational costs in Large Language Models (LLMs). By reducing the number of generated "think tokens" by up to 90\% compared to standard reasoning methods, AdaptR1 significantly lowers the energy consumption and carbon footprint associated with model inference. This aligns with the goals of Green AI.

However, we acknowledge that our method relies on the pre-trained Qwen2.5-7B-Instruct model. Like all LLMs, this backbone model may carry inherent biases or the potential to generate toxic content derived from its training data. While our adaptive strategy aims to improve efficiency and does not explicitly introduce new biases, it does not actively mitigate existing ones. Users should exercise caution and implement appropriate safety guardrails when deploying such models in real-world applications.

AI assistants were used for language polishing, improving clarity and readability, and editorial integration of author-provided revision material. All scientific content, including the research ideas, methodology, experimental design, results, and conclusions, was conceived, implemented, and verified by the authors. The use of AI tools did not influence the technical decisions or the interpretation of experimental results.

\section{GRPO Objective}
\label{sec:appendix-grpo}
For each question $q \sim P(Q)$, GRPO samples a group of outputs $\{o_1, o_2, \dots, o_G\}$ from the old policy $\pi_{\theta_{\text{old}}}$ and updates the policy model $\pi_{\theta}$ by optimizing:
\begin{equation}
\begin{aligned}
\mathcal{J}_{\text{GRPO}}(\theta)
= \mathbb{E}_{q \sim P(Q),\, \{o_i\}_{i=1}^G \sim \pi_{\theta_{\text{old}}}(O|q)} \\
\Bigg[
\frac{1}{G} \sum_{i=1}^G \frac{1}{|o_i|} \sum_{t=1}^{|o_i|}
\Big(
\min\big(r_t(\theta)\hat{A}_{i,t},\\
\, \operatorname{clip}(r_t(\theta),\, 1-\varepsilon,\, 1+\varepsilon)\hat{A}_{i,t}\big) \\
- \beta\, \mathbb{D}_{\text{KL}}(\pi_{\theta} \,||\, \pi_{\text{ref}})
\Big)
\Bigg].
\end{aligned}
\label{eq:grpo}
\end{equation}
Here, $r_t(\theta)=\frac{\pi_\theta(o_{i,t} \mid q, o_{i,<t})}{\pi_{\theta_{\text{old}}}(o_{i,t} \mid q, o_{i,<t})}$ is the probability ratio, $\varepsilon$ controls clipping, $\beta$ regulates the KL penalty, and $\hat{A}_{i,t}$ is computed using group-relative reward:
\begin{equation}
\hat{A}_{i,t} = \frac{r_i - \mathrm{mean}(\mathbf{r})}{\mathrm{std}(\mathbf{r})}.
\label{eq:advantage}
\end{equation}
where $\mathbf{r} = \{r_1, r_2, \dots, r_G\}$ is the reward vector of $G$ outputs. Since GRPO uses sequence-level rewards, $\hat{A}_{i,t}$ is constant across all tokens within the same trajectory.

\section{Algorithm}
\label{sec:appendix-algo}
We show the algorithm for AdaptR1.
\begin{algorithm*}[htbp]
\caption{AdaptR1}
\label{alg:llm_search}
\begin{algorithmic}[1]
\Require Input \( x \), LLM \( \pi_{\theta} \), Retrieval set \( \mathcal{R} \), Max turns \( B \).
\Ensure Output \( y \).

\State Initialize \( y \gets \emptyset \)
% \State Initialize search call count \( b \gets 0 \)
\State Initialize \( b \gets 0 \)
\While{\( b < B \)}
    \State Rollout   \( y_b \gets \emptyset \) 
    \While{True}
    \State \textbf{Adaptively} generate either a reasoning trace or \texttt{<think>no\_think</think>}: \( y_t \sim \pi_{\theta}(\cdot \mid x, y + y_b) \)
    \State concatenate token   \( y_b \gets y_b + y_t \)
    \If{\( y_t \) in [\textcolor{orange}{\texttt{</query>}}, \textcolor{purple}{\texttt{</answer>}}, \texttt{<eos>}]}
        break
    \EndIf
    \EndWhile

    \State \( y  \gets  y + y_b \)
    \If{extract \textcolor{orange}{\texttt{<query>}} \textcolor{orange}{\texttt{</query>}} from \( y_b \)}
        \State Extract \( q \gets \text{Parse}(y_b,\textcolor{orange}{\texttt{<query>}}, \textcolor{orange}{\texttt{</query>}} ) \)
        \State Retrieve knowledge \( d = \mathcal{R}(q) \)
        \State Continue rollout \( y  \gets  y + \textcolor{teal}{\texttt{<knowledge>}}d\textcolor{teal}{\texttt{</knowledge>}}  \)
    \ElsIf{extract \textcolor{purple}{\texttt{</answer>}} from \( y_b \)}
        % \State // \textcolor{cvprblue}{Terminate rollout}
        \State \textbf{return} \( y \)
    \EndIf

    \State count turns \( b \gets b + 1 \)
\EndWhile

\State \textbf{return} \( y \)
\end{algorithmic}
\end{algorithm*}

\section{Additional Experiments}
\subsection{Robustness Across Random Seeds}
\label{appendix:robustness}
We further evaluate robustness with 5 random seeds on three representative datasets. Table~\ref{tab:robustness-seeds} reports mean and standard deviation for answer metrics and think-token usage. Accuracy variance remains moderate, and the average think-token count remains substantially below the corresponding baseline in every dataset.

\begin{table*}[htbp]
\centering
\small
\begin{tabular}{llcc}
\toprule
\textbf{Dataset} & \textbf{Metric} & \textbf{Baseline} & \textbf{AdaptR1 (5 Seeds)} \\
\midrule
\multirow{3}{*}{2WikiMultiHopQA} & EM & 58.59 & $61.72 \pm 1.48$ \\
 & F1 & 68.18 & $68.73 \pm 1.50$ \\
 & Avg. Think Tokens & 151.14 & $25.43 \pm 18.81$ \\
\midrule
\multirow{3}{*}{HotpotQA} & EM & 55.47 & $60.31 \pm 1.76$ \\
 & F1 & 63.55 & $65.93 \pm 1.02$ \\
 & Avg. Think Tokens & 103.68 & $13.81 \pm 2.22$ \\
\midrule
\multirow{3}{*}{Musique} & EM & 37.50 & $39.69 \pm 1.28$ \\
 & F1 & 48.33 & $50.39 \pm 2.21$ \\
 & Avg. Think Tokens & 148.29 & $45.69 \pm 36.88$ \\
\bottomrule
\end{tabular}
\caption{Robustness statistics over 5 random seeds on three representative datasets.}
\label{tab:robustness-seeds}
\end{table*}

\subsection{No-Think Ratios}
\label{appendix:nothinkratios}
We show the No-Think ratios of left 5 datasets in Table \ref{tab:ablation_step-wise-token-saving-2wikimultihopqa},\ref{tab:ablation_step-wise-token-saving-hotpotqa},\ref{tab:ablation_step-wise-token-saving-nq},\ref{tab:ablation_step-wise-token-saving-popqa},\ref{tab:ablation_step-wise-token-saving-triviaqa}.
\begin{table*}[htbp]
    \centering
    \small
    \begin{tabular}{l|cccccc|cl}
    \toprule
             &\textbf{Step 1} & \textbf{Step 2} & \textbf{Step 3} & \textbf{Step 4} & \textbf{Step 5} & \textbf{Step 6} & \textbf{Avg. Ratio} &  \textbf{Avg. F1}\\
        \midrule
        \multicolumn{9}{l}{\textit{Impact of Step-wise Weight ($\lambda$)}}\\
        \midrule
        $\lambda=0.5$ & 0.0000 & 0.0000 & 0.0000 & 0.0000 & 0.0000 & - & 0.0000 & 0.6827\\
        \midrule
        \multicolumn{9}{l}{\textit{Sensitivity Analysis}}\\
        $\lambda=0.1$ & 0.0000 & 0.0000 & 0.5000 & 0.8966 & 0.9524 & 0.9524 & 0.4408 & 0.6819\\
        $\lambda=0.2$ & 0.0000 & 0.0000 & 0.3047 & 0.8440 & 0.7742 & 1.0000 & 0.3025 & 0.6756\\
        $\lambda=0.3$ & 0.0000 & 0.0000 & 0.0000 & 0.0000 & 0.0000 & 0.0000 & 0.0000 & 0.6870\\
        $\lambda=0.4$ & 0.4219 & 0.4297 & 1.0000 & 1.0000 & 1.0000 & - & 0.7184 & 0.6599\\
        \midrule
        $\lambda=0.6$ & 0.0000 & 0.0000 & 0.0000 & 0.0000 & - & - & 0.0000 & 0.6959\\
        $\lambda=0.7$ & 1.0000 & 0.2344 & 1.0000 & - & - & - & 0.7216 & 0.6370\\
        $\lambda=0.8$ & 0.0312 & 0.0000 & 0.0000 & 0.0000 & 0.0000 & - & 0.0102 & 0.6572\\
        $\lambda=0.9$ & 1.0000 & 0.0000 & 0.9921 & 0.2381 & 0.0000 & - & 0.6174 & 0.6920\\
        \bottomrule
    \end{tabular}
    \caption{Analysis of the No-Think ratio from step 1 to 6 and performance (F1) for Graph-R1 with and without AdaptR1 on 2WikiMultiHopQA. The parameter $\lambda$ controls the step-wise penalty weight.}
    \label{tab:ablation_step-wise-token-saving-2wikimultihopqa}
\end{table*}

\begin{table*}[htbp]
    \centering
    \small
    \begin{tabular}{l|cccccc|cl}
    \toprule
             &\textbf{Step 1} & \textbf{Step 2} & \textbf{Step 3} & \textbf{Step 4} & \textbf{Step 5} & \textbf{Step 6} & \textbf{Avg. Ratio} &  \textbf{Avg. F1}\\
        \midrule
        \multicolumn{9}{l}{\textit{Impact of Step-wise Weight ($\lambda$)}}\\
        \midrule
        $\lambda=0.5$ & 1.0000 & 1.0000 & 1.0000 & - & - & - & 1.0000 & 0.6081\\
        \midrule
        \multicolumn{9}{l}{\textit{Sensitivity Analysis}}\\
        $\lambda=0.1$ & 1.0000 & 1.0000 & - & - & - & - & 1.0000 & 0.6272\\
        $\lambda=0.2$ & 1.0000 & 1.0000 & 1.0000 & 1.0000 & 1.0000 & 1.0000 & 0.9972 & 0.6460\\
        $\lambda=0.3$ & 1.0000 & 1.0000 & - & - & - & - & 1.0000 & 0.6421\\
        $\lambda=0.4$ & 1.0000 & 1.0000 & 1.0000 & - & - & - & 1.0000 & 0.6109\\
        \midrule
        $\lambda=0.6$ & 1.0000 & 1.0000 & 1.0000 & - & - & - & 1.0000 & 0.6205\\
        $\lambda=0.7$ & 1.0000 & 1.0000 & 1.0000 & - & - & - & 1.0000 & 0.6240\\
        $\lambda=0.8$ & 1.0000 & 1.0000 & 1.0000 & 1.0000 & 1.0000 & 0.9167 & 0.9924 & 0.6483\\
        $\lambda=0.9$ & 1.0000 & 1.0000 & - & - & - & - & 1.0000 & 0.6211\\
        \bottomrule
    \end{tabular}
    \caption{Analysis of the No-Think ratio from step 1 to 6 and performance (F1) for Graph-R1 with and without AdaptR1 on HotpotQA. The parameter $\lambda$ controls the step-wise penalty weight.}
    \label{tab:ablation_step-wise-token-saving-hotpotqa}
\end{table*}

\begin{table*}[htbp]
    \centering
    \small
    \begin{tabular}{l|cccccc|cl}
    \toprule
             &\textbf{Step 1} & \textbf{Step 2} & \textbf{Step 3} & \textbf{Step 4} & \textbf{Step 5} & \textbf{Step 6} & \textbf{Avg. Ratio} &  \textbf{Avg. F1}\\
        \midrule
        \multicolumn{9}{l}{\textit{Impact of Step-wise Weight ($\lambda$)}}\\
        \midrule
        $\lambda=0.5$ & 1.0000 & 1.0000 & 1.0000 & 1.0000 & - & - & 1.0000 & 0.4686\\
        \midrule
        \multicolumn{9}{l}{\textit{Sensitivity Analysis}}\\
        $\lambda=0.1$ & 0.9922 & 0.0156 & 0.0000 & 0.0000 & 0.0000 & - & 0.4868 & 0.4704\\
        $\lambda=0.2$ & 1.0000 & 1.0000 & - & - & - & - & 1.0000 & 0.4641\\
        $\lambda=0.3$ & 1.0000 & 1.0000 & - & - & - & - & 1.0000 & 0.4623\\
        $\lambda=0.4$ & 1.0000 & 1.0000 & 1.0000 & - & - & - & 1.0000 & 0.4875\\
        \midrule
        $\lambda=0.6$ & 1.0000 & 1.0000 & - & - & - & - & 1.0000 & 0.4842\\
        $\lambda=0.7$ & 1.0000 & 1.0000 & - & - & - & - & 1.0000 & 0.4937\\
        $\lambda=0.8$ & 1.0000 & 1.0000 & - & - & - & - & 1.0000 & 0.4729\\
        $\lambda=0.9$ & 1.0000 & 1.0000 & - & - & - & - & 1.0000 & 0.4914\\
        \bottomrule
    \end{tabular}
    \caption{Analysis of the No-Think ratio from step 1 to 6 and performance (F1) for Graph-R1 with and without AdaptR1 on NQ. The parameter $\lambda$ controls the step-wise penalty weight. }
    \label{tab:ablation_step-wise-token-saving-nq}
\end{table*}

\begin{table*}[htbp]
    \centering
    \small
    \begin{tabular}{l|cccccc|cl}
    \toprule
             &\textbf{Step 1} & \textbf{Step 2} & \textbf{Step 3} & \textbf{Step 4} & \textbf{Step 5} & \textbf{Step 6} & \textbf{Avg. Ratio} &  \textbf{Avg. F1}\\
        \midrule
        \multicolumn{9}{l}{\textit{Impact of Step-wise Weight ($\lambda$)}}\\
        \midrule
        $\lambda=0.5$ & 1.0000 & 1.0000 & - & - & - & - & 1.0000 & 0.5573\\
        \midrule
        \multicolumn{9}{l}{\textit{Sensitivity Analysis}}\\
        $\lambda=0.1$ & 1.0000 & 1.0000 & - & - & - & - & 1.0000 & 0.5584\\
        $\lambda=0.2$ & 1.0000 & 1.0000 & 1.0000 & 1.0000 & 1.0000 & 1.0000 & 1.0000 & 0.5495\\
        $\lambda=0.3$ & 1.0000 & 1.0000 & - & - & - & - & 1.0000 & 0.5536\\
        $\lambda=0.4$ & 1.0000 & 1.0000 & - & - & - & - & 1.0000 & 0.5573\\
        \midrule
        $\lambda=0.6$ & 1.0000 & 1.0000 & - & - & - & - & 1.0000 & 0.5589\\
        $\lambda=0.7$ & 1.0000 & 1.0000 & 1.0000 & 1.0000 & 1.0000 & - & 1.0000 & 0.5539\\
        $\lambda=0.8$ & 1.0000 & 1.0000 & - & - & - & - & 1.0000 & 0.5683\\
        $\lambda=0.9$ & 1.0000 & 1.0000 & - & - & - & - & 1.0000 & 0.5284\\
        \bottomrule
    \end{tabular}
    \caption{Analysis of the No-Think ratio from step 1 to 6 and performance (F1) for Graph-R1 with and without AdaptR1 on PopQA. The parameter $\lambda$ controls the step-wise penalty weight. }
    \label{tab:ablation_step-wise-token-saving-popqa}
\end{table*}

\begin{table*}[htbp]
    \centering
    \small
    \begin{tabular}{l|cccccc|cl}
    \toprule
             &\textbf{Step 1} & \textbf{Step 2} & \textbf{Step 3} & \textbf{Step 4} & \textbf{Step 5} & \textbf{Step 6} & \textbf{Avg. Ratio} &  \textbf{Avg. F1}\\
        \midrule
        \multicolumn{9}{l}{\textit{Impact of Step-wise Weight ($\lambda$)}}\\
        \midrule
        $\lambda=0.5$ & 1.0000 & 1.0000 & - & - & - & - & 1.0000 & 0.7319\\
        \midrule
        \multicolumn{9}{l}{\textit{Sensitivity Analysis}}\\
        $\lambda=0.1$ & 1.0000 & 1.0000 & - & - & - & - & 1.0000 & 0.7157\\
        $\lambda=0.2$ & 0.9766 & 1.0000 & - & - & - & - & 0.9883 & 0.7290\\
        $\lambda=0.3$ & 1.0000 & 1.0000 & 1.0000 & - & - & - & 1.0000 & 0.7156\\
        $\lambda=0.4$ & 1.0000 & 1.0000 & - & - & - & - & 1.0000 & 0.7279\\
        \midrule
        $\lambda=0.6$ & 1.0000 & 1.0000 & - & - & - & - & 1.0000 & 0.7213\\
        $\lambda=0.7$ & 1.0000 & 1.0000 & - & - & - & - & 1.0000 & 0.7100\\
        $\lambda=0.8$ & 1.0000 & 0.9531 & - & - & - & - & 0.9766 & 0.7181\\
        $\lambda=0.9$ & 1.0000 & 0.9922 & - & - & - & - & 0.9961 & 0.7277\\
        \bottomrule
    \end{tabular}
    \caption{Analysis of the No-Think ratio from step 1 to 6 and performance (F1) for Graph-R1 with and without AdaptR1 on TriviaQA. The parameter $\lambda$ controls the step-wise penalty weight. }
    \label{tab:ablation_step-wise-token-saving-triviaqa}
\end{table*}

\subsection{Training Dynamics}
\label{appendix:trainingdynamics}
\begin{figure*}
    \centering
    \includegraphics[width=\linewidth]{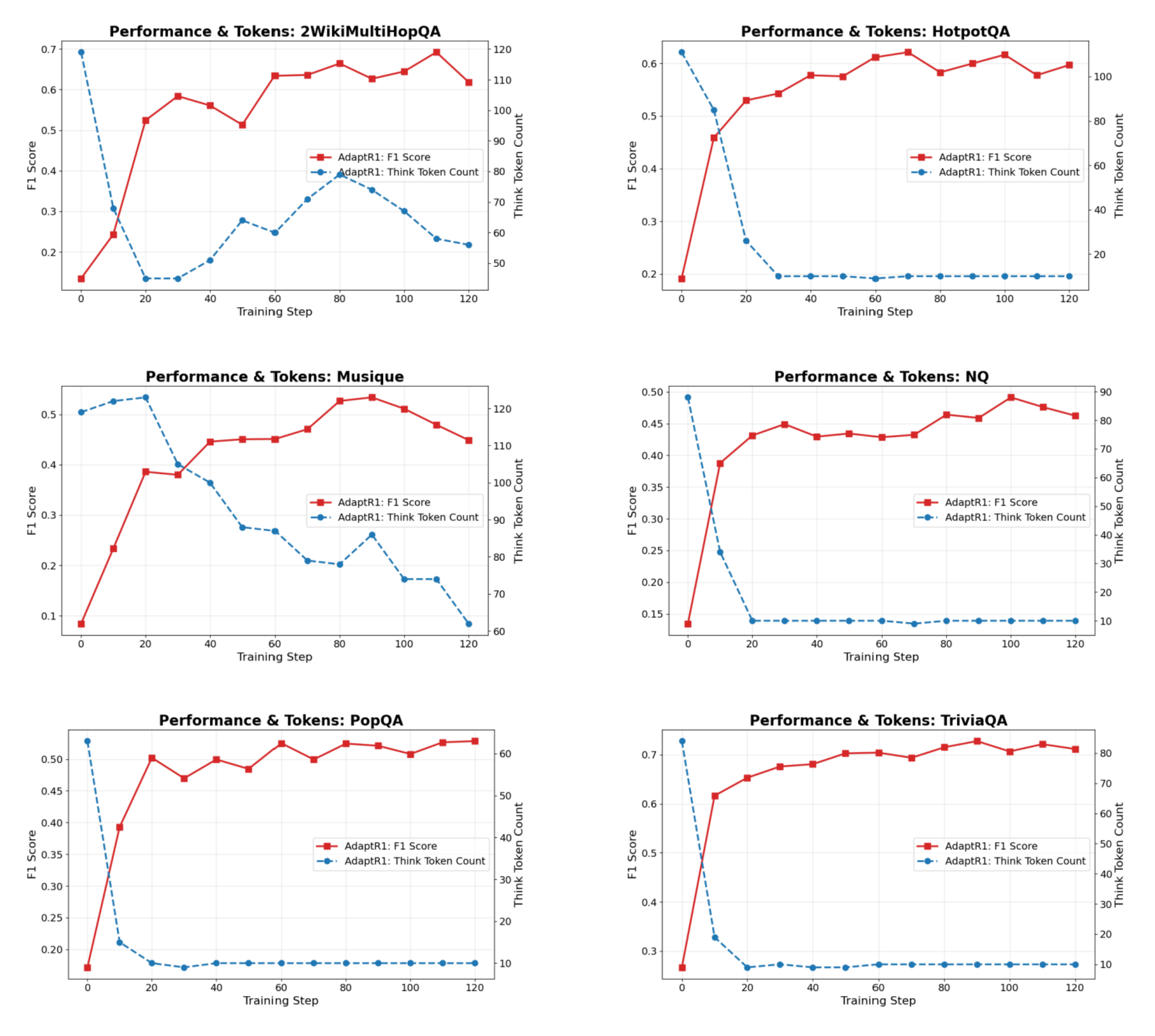}
    \caption{Think tokens and F1 scores in the training steps for six datasets.}
    \label{fig:trainingdynamicsall}
\end{figure*}

We show the training dynamics of six datasets in Figure \ref{fig:trainingdynamicsall}.

\subsection{Case Study}
\label{appendix:casestudy}
We show the case of Graph-R1 in Table \ref{tab:case-graphr1} and Graph-AdaptR1 in Table \ref{tab:case-graphAdapt-R1}.

\begin{table*}[h]
    \centering
    \footnotesize
    \begin{tabular}{p{13.5cm}}
        \hline
        \textbf{Question}: What is the birthplace of the man who does the voice of Stan on the series that includes the episode The Hobbit?\\
        \hline
        \textbf{Ground Truth}: "Denver" or 
            "Denver, Colorado"\\
        \hline
        \textbf{Graph-R1}: \\
        \think{To answer this question, I need to find the birthplace of the voice actor who does the voice of Stan in the series that includes the episode "The Hobbit". First, I will determine the name of the series that includes the episode "The Hobbit". Then, I will find the birthplace of the actor who voices Stan in that series.} \\
        \search{\{"query": "what series includes the episode The Hobbit?"\}}\\
        \info{\{"results": [{"<knowledge>": "The Hobbit is the tenth episode in the seventeenth season of 'South Park', with a storyline centered on Wendy Testaburger's attempts at body image awareness.", "<coherence>": 2.333}, {"<knowledge>": "The Hobbit is the tenth and final episode in the seventeenth season of the American animated television series South Park. The 247th episode of the series overall", "<coherence>": 1.283}, {"<knowledge>": "It was commissioned for the soundtrack of the 2013 film 'The Hobbit', where it was played over the closing credits.", "<coherence>": 0.667}, {"<knowledge>": "Kidnapped is a two-part BBC television adaptation of the 1886 novel of the same name by Robert Louis Stevenson.", "<coherence>": 0.25}, {"<knowledge>": "'The Hobbit' director, Peter Jackson, asked Sheeran to write a song for the movie after Jackson's daughter, Katie, suggested Sheeran.", "<coherence>": 0.2}, {"<knowledge>": "The episode received positive reviews, praising Wendy's story arc and the return of Kanye West.", "<coherence>": 0.143}, {"<knowledge>": "The episode is noted as the 247th episode of 'South Park', showcasing the series' long run and significance.", "<coherence>": 0.125}]}\}\\
        \think{Based on the knowledge base, the series that includes the episode "The Hobbit" is South Park. Now, I need to find the birthplace of the actor who voices Stan in South Park.}\\
        \search{\{"query": "birthplace of the actor who voices Stan in South Park"\}}\\
        \info{\{"results": [{"<knowledge>": "Trey Parker voices several main characters including Stan\ Marsh, Eric Cartman, Randy Marsh, and Mr. Garrison, as well as recurring characters in 'South Park.'", "<coherence>": 3.283}, {"<knowledge>": "Dian Bachar (; born October 26, 1970, in Denver, Colorado) is an American actor most notable for his roles in various films by or starring his friends Trey Parker and Matt Stone, such as Cannibal! The Musical (George Noon), Orgazmo (Ben Chapleski) and his most famous role as Kenny Squeak Scolari in 1998's BASEketball, as well as making the occasional appearance on South Park.", "<coherence>": 0.5}, {"<knowledge>": "Big Stan is a 2007 American prison comedy directed and produced by Rob Schneider, who also starred in the film.", "<coherence>": 0.333}, {"<knowledge>": "Kenny 'Squeak' Scolari was known for his role in 1998's 'BASEketball', along with appearances on 'South Park' and portraying an alien engineer in 'Galaxy Quest'.", "<coherence>": 0.25}, {"<knowledge>": "He was raised in Collingwood, Ontario.", "<coherence>": 0.2}]\}}\\
        \think{Based on the knowledge base, Trey Parker voices Stan Marsh in South Park. Therefore, I need to find the birthplace of Trey Parker.}\\
        \search{\{"query": "birthplace of Trey Parker"}\}\\
        \info{\{"results": [\{"<knowledge>": "Henry Benbridge (October 1743 u2013 February 1812) was an early American portrait painter.", "<coherence>": 1.0\}, \{"<knowledge>": "Born in Enugu, Nigeria on March 1, 1980, Anthony Njokuani and his family immigrated to the United States in 1983, settling in Garland, Texas.", "<coherence>": 1.0\}, \{"<knowledge>": "Dian Bachar (; born October 26, 1970, in Denver, Colorado) is an American actor most notable for his roles in various films by or starring his friends Trey Parker and Matt Stone, such as Cannibal! The Musical (George Noon), Orgazmo (Ben Chapleski) and his most famous role as Kenny Squeak Scolari in 1998's BASEketball, as well as making the occasional appearance on South Park.", "<coherence>": 0.553\}, \{"<knowledge>": "u201cAre You Ready for Love" is a song recorded by Elton John in 1977 and first released in the UK in 1979 on the EP "The Thom Bell Sessions". It was written by Leroy Bell, Thom Bell and Casey James, and was originally produced in Philadelphia by Thom Bell.", "<coherence>": 0.5\}, \{"<knowledge>": "Trey Parker voices several main characters including Stan Marsh, Eric Cartman, Randy Marsh, and Mr. Garrison, as well as recurring characters in 'South Park.'", "<coherence>": 0.383\}, \{"<knowledge>": "The Federal Detention Center (FDC Philadelphia) is a United States Federal prison in Center City, Philadelphia, Pennsylvania.", "<coherence>": 0.333\}, \{"<knowledge>": "The development of the appreciation medals was initiated by the White House and executed by the Bureau of the Mint through the U.S. Mint in Philadelphia.", "<coherence>": 0.25\}, \{"<knowledge>": "United States. It was the birthplace and boyhood home of Dr. George Horace Gallup, the pioneer of public opinion polling.", "<coherence>": 0.25\}, \{"<knowledge>": "He was born in Philadelphia, Pennsylvania, the eldest son of a well-known Quaker family (Richard and Anne Roberts Vaux) and connected by marriage to another such family, the Wistars.", "<coherence>": 0.2\}, \{"<knowledge>": "Nathaniel Carlyle Adderley was born in Tampa, Florida, but moved to Tallahassee when his parents were hired to teach at Florida A\&M University.", "<coherence>": 0.2\}]\}}\\
        \think{Based on the knowledge base, Trey Parker was born in Philadelphia, Pennsylvania.}\\
        \answer{Philadelphia, Pennsylvania} \\
        \hline
    \end{tabular}
    \caption{A case study of Graph-R1.}\label{tab:case-graphr1}
\end{table*}

\begin{table*}[h]
    \centering
    \footnotesize
    \begin{tabular}{p{13.5cm}}
        \hline
        \textbf{Question}: What is the birthplace of the man who does the voice of Stan on the series that includes the episode The Hobbit?\\
        \hline
        \textbf{Ground Truth}: "Denver" or 
            "Denver, Colorado"\\
        \hline
        \textbf{Graph-AdaptR1}: \\
        \think{no\_think}\\
        \search{\{"query": "What is the birthplace of the man who does the voice of Stan on the series that includes the episode The Hobbit?"\}}\\
        \info{\{"results": [\{"<knowledge>": "Trey Parker voices several main characters including Stan Marsh, Eric Cartman, Randy Marsh, and Mr. Garrison, as well as recurring characters in 'South Park.'", "<coherence>": 1.5\}, \{"<knowledge>": "with 'Stan Cohen'."<", "<coherence>": 1.0\}, \{"<knowledge>": "The film co-stars Jennifer Morrison, Scott Wilson and David Carradine.", "<coherence>": 0.583\}, \{"<knowledge>": "The Lorax (also known as Dr. Seuss' The Lorax) is a 2012 American 3D computer-animated musical fantasy-comedy film produced by Illumination Entertainment and based on Dr. Seuss's children's book of the same name.", "<coherence>": 0.5\}, \{"<knowledge>": "The Hobbit is the tenth and final episode in the seventeenth season of the American animated television series South Park. The 247th episode of the series overall", "<coherence>": 0.4\}, \{"<knowledge>": "The Hobbit is the tenth episode in the seventeenth season of 'South Park', with a storyline centered on Wendy Testaburger's attempts at body image awareness.", "<coherence>": 0.333\}, \{"<knowledge>": "Dian Bachar (; born October 26, 1970, in Denver, Colorado) is an American actor most notable for his roles in various films by or starring his friends Trey Parker and Matt Stone, such as Cannibal! The Musical (George Noon), Orgazmo (Ben Chapleski) and his most famous role as Kenny Squeak Scolari in 1998's BASEketball, as well as making the occasional appearance on South Park.", "<coherence>": 0.25\}]\}}\\
        \think{The man who does the voice of Stan on the series that includes the episode "The Hobbit" is Trey Parker. The birthplace of Trey Parker is Denver, Colorado.}\\
        \search{\{"query": "What is the birthplace of Trey Parker?"}\}\\
        \info{\{"results": [\{"<knowledge>": "Henry Benbridge (October 1743 u2013 February 1812) was an early American portrait painter.", "<coherence>": 1.0\}, \{"<knowledge>": "Trey Parker voices several main characters including Stan Marsh, Eric Cartman, Randy Marsh, and Mr. Garrison, as well as recurring characters in 'South Park.'", "<coherence>": 1.0\}, \{"<knowledge>": "u201cAre You Ready for Love" is a song recorded by Elton John in 1977 and first released in the UK in 1979 on the EP "The Thom Bell Sessions". It was written by Leroy Bell, Thom Bell and Casey James, and was originally produced in Philadelphia by Thom Bell.", "<coherence>": 0.5\}, \{"<knowledge>": "Dian Bachar (; born October 26, 1970, in Denver, Colorado) is an American actor most notable for his roles in various films by or starring his friends Trey Parker and Matt Stone, such as Cannibal! The Musical (George Noon), Orgazmo (Ben Chapleski) and his most famous role as Kenny Squeak Scolari in 1998's BASEketball, as well as making the occasional appearance on South Park.", "<coherence>": 0.5\}, \{"<knowledge>": "The Federal Detention Center (FDC Philadelphia) is a United States Federal prison in Center City, Philadelphia, Pennsylvania.", "<coherence>": 0.333\}, \{"<knowledge>": "United States. It was the birthplace and boyhood home of Dr. George Horace Gallup, the pioneer of public opinion polling.", "<coherence>": 0.333\}, \{"<knowledge>": "The development of the appreciation medals was initiated by the White House and executed by the Bureau of the Mint through the U.S. Mint in Philadelphia.", "<coherence>": 0.25\}, \{"<knowledge>": "Nathaniel Carlyle Adderley was born in Tampa, Florida, but moved to Tallahassee when his parents were hired to teach at Florida A\&M University.", "<coherence>": 0.25\}, \{"<knowledge>": "He was born in Philadelphia, Pennsylvania, the eldest son of a well-known Quaker family (Richard and Anne Roberts Vaux) and connected by marriage to another such family, the Wistars.", "<coherence>": 0.2\}, \{"<knowledge>": "Cassavetes was born in New York City, the son of Greek American Katherine Cassavetes and Greek immigrant Nicholas John Cassavetes.", "<coherence>": 0.2\}]\}}\\
        \think{no\_think}\\
        \answer{Denver, Colorado}\\
        \hline
    \end{tabular}
    \caption{A case study of Graph-AdaptR1.}\label{tab:case-graphAdapt-R1}
\end{table*}

\subsection{Failure Analysis}
\label{appendix:failureanalysis}
We analyze when \texttt{no\_think} can hurt answer quality by comparing Graph-AdaptR1 with the full-thinking Graph-R1 baseline. Table~\ref{tab:failure-overlap} separates examples that only AdaptR1 answers correctly, examples both methods answer correctly, and cases where Graph-R1 is correct but Graph-AdaptR1 fails. The last category approximates failures where explicit reasoning was likely useful but was skipped or shortened too aggressively.

\begin{table*}[htbp]
\centering
\small
\begin{tabular}{lccc}
\toprule
\textbf{Dataset} & \textbf{Only AdaptR1 Correct} & \textbf{Both Correct} & \textbf{Only Graph-R1 Correct} \\
\midrule
2WikiMultiHopQA & 15 & 64 & 11 \\
HotpotQA & 14 & 60 & 11 \\
Musique & 13 & 38 & 10 \\
NQ & 3 & 43 & 2 \\
PopQA & 7 & 56 & 9 \\
TriviaQA & 4 & 78 & 7 \\
\bottomrule
\end{tabular}
\caption{Overlap analysis between Graph-AdaptR1 and Graph-R1. ``Only Graph-R1 Correct'' indicates cases where skipping or shortening explicit reasoning likely harms the answer.}
\label{tab:failure-overlap}
\end{table*}

The failures are relatively infrequent, and in all datasets except NQ the number of examples fixed by AdaptR1 is larger than the number lost by AdaptR1. Qualitatively, the main failure mode appears in strict comparison questions that require precise extraction and comparison of attributes, such as dates or numerical values. For example, for the question ``Which film has the director who died later, \textit{The Hellions} or \textit{Hum Kaun Hai}?'' Graph-AdaptR1 answers \textit{The Hellions} after skipping final synthesis, while Graph-R1 retrieves the directors separately, compares their death years, and answers \textit{Hum Kaun Hai}. This suggests that \texttt{no\_think} is most risky when the final step requires explicit symbolic comparison over retrieved evidence.

\section{Experimental Settings}
\label{appendix:detailedimplementation}

\subsection{Baselines in the Graph-R1 Setting}

We categorize our baselines into two distinct groups based on the underlying backbone model.

\noindent\textbf{GPT-4o-mini Based Methods.} 
The first group employs \textbf{GPT-4o-mini} as an inference-only generator. We evaluate the base model's intrinsic capacity using \textbf{NaiveGeneration}, a zero-shot approach without retrieval. We also include \textbf{StandardRAG} \citep{RAG}, representing the conventional chunk-based retrieval-augmented generation paradigm. Furthermore, we assess a suite of graph-based retrieval strategies: \textbf{GraphRAG} \citep{GraphRAG}, which constructs entity graphs for one-shot retrieval; \textbf{LightRAG} \citep{LightRAG}, a streamlined variant designing compact graphs for efficiency; \textbf{PathRAG} \citep{PathRAG}, which executes retrieval via path-based pruning on entity graphs; \textbf{HippoRAG2} \citep{HippoRAG2}, utilizing a hierarchical path planner over knowledge graphs; and \textbf{HyperGraphRAG} \citep{HyperGraphRAG}, which leverages n-ary relational hypergraphs to facilitate single-step retrieval.

\noindent\textbf{Qwen2.5-Instruct Based Methods.}
The second group utilizes the \textbf{\texttt{Qwen2.5-Instruct} (7B)} model. We establish foundational performance bounds using \textbf{NaiveGeneration}, the classic \textbf{StandardRAG} \citep{RAG} pipeline, and \textbf{SFT} \citep{SFT}, which applies supervised fine-tuning on QA pairs. Additionally, we evaluate advanced methods optimized via reinforcement learning (RL): \textbf{R1} \citep{GRPO}, a policy trained with GRPO to generate answers directly without retrieval; \textbf{Search-R1} \citep{Search-R1}, a multi-turn chunk-based retrieval approach trained via GRPO; \textbf{R1-Searcher} \citep{R1-Searcher}, a two-stage GRPO-based framework for chunk-based retrieval; and \textbf{Graph-R1} \citep{Graph-R1}, an agentic GraphRAG framework enhanced by end-to-end reinforcement learning.

\subsection{Evaluation Metrics}

We assess model performance using three primary metrics focusing on answer accuracy and retrieval quality.

\noindent\textbf{Exact Match (EM).}
We employ Exact Match to strictly evaluate answer accuracy. This metric determines whether the generated answer $y_i$ is identical to the ground-truth reference $y_i^\star$ following a normalization process (i.e., lowercasing, punctuation removal, and whitespace standardization). The EM score is averaged over all $N$ samples:
\begin{equation}
  \text{EM} = \frac{1}{N} \sum_{i=1}^{N} \mathbb{I} \left[ \text{norm}(y_i) = \text{norm}(y_i^\star) \right],
\end{equation}
where $\mathbb{I}[\cdot]$ denotes the indicator function.

\noindent\textbf{F1 Score.} 
To provide a more granular assessment of generation quality beyond binary matching, we utilize the F1 score. This metric measures the token-level overlap between the prediction and the ground truth, defined as the harmonic mean of precision and recall:
\begin{equation}
  \text{F1} = \frac{1}{N} \sum_{i=1}^{N} \frac{2 \cdot |\mathcal{T}(y_i) \cap \mathcal{T}(y_i^\star)|}{|\mathcal{T}(y_i)| + |\mathcal{T}(y_i^\star)|},
\end{equation}
where $\mathcal{T}(\cdot)$ represents the set of tokens in a given text.

\noindent\textbf{Retrieval Similarity (R-S).}
To assess the efficacy of the retrieval module in isolation, we compute Retrieval Similarity. This metric measures the semantic alignment between the retrieved context $k_{\text{retr}}^{(i)}$ and the ground-truth "gold" context $k_{\text{gold}}^{(i)}$. We utilize a semantic embedding function, $\text{Enc}(\cdot)$, to compute the cosine similarity between the vector representations:
\begin{equation}
  \text{R-S} = \frac{1}{N} \sum_{i=1}^{N} \cos\left(\text{Enc}(k_{\text{retr}}^{(i)}), \text{Enc}(k_{\text{gold}}^{(i)})\right).
\end{equation}

\subsection{Implementation Details}
We outline the general hyperparameters in Table~\ref{tab:hyperparams}. While our AdaptR1 models largely inherit the configurations from their respective backbones, we introduce specific adjustments to the reinforcement learning setup to facilitate adaptive training. We configure the group size $G$ (number of rollouts per query) to 8 for Graph-AdaptR1 and 5 for Search-AdaptR1. For Search-AdaptR1, we add the step-wise No-Think reward to the original answer and retrieval set rewards. For the AdaptR1-specific coefficients, we set the confidence threshold $\tau=0.6$, the reward weight $\omega=0.2$, and the step-wise penalty factor $\lambda=0.9$. Main and ablation experiments were conducted on NVIDIA H200 GPUs; robustness statistics in Table~\ref{tab:robustness-seeds} are repeated over 5 random seeds on three representative datasets. The total computational budget for all reported experiments (across all datasets and ablation studies) was approximately 1500 GPU hours.

\begin{table*}[htbp]
\centering
\fontsize{7pt}{7.5pt}\selectfont
\setlength{\tabcolsep}{2.8mm}{
\begin{tabular}{lcccccc}
\toprule
\textbf{Method} & \textbf{Backbone} & \textbf{Batch Size} & \textbf{Max Length} & \textbf{Top-K} & \textbf{Algo} & \textbf{Epochs} \\
\midrule
NaiveGeneration     & Qwen2.5 / GPT-4o-mini & --   & $\infty$ & N/A  & --     & --  \\
StandardRAG         & Qwen2.5 / GPT-4o-mini & --   & $\infty$ & 5 Chunks  & --     & --  \\
GraphRAG            & GPT-4o-mini           & --   & $\infty$ & 60 & --     & --  \\
LightRAG            & GPT-4o-mini           & --   & $\infty$ & 60  & --     & --  \\
PathRAG             & GPT-4o-mini           & --   & $\infty$ & 60  & --     & --  \\
HippoRAG2           & GPT-4o-mini           & --   & $\infty$ & 60  & --     & --  \\
HyperGraphRAG       & GPT-4o-mini           & --   & $\infty$ & 60  &  --     & --  \\
SFT                 & Qwen2.5 (7B)              & 16   & 4096 & N/A  & LoRA     & 3   \\
 R1                  & Qwen2.5 (7B)               & 128   & 4096 & N/A  & GRPO   & 3   \\
 R1-Searcher         & Qwen2.5 (7B)               & 128   & 4096 & 5 Chunks / Turn & GRPO & 3 \\
\rowcolor{R1!15} Search-R1           & Qwen2.5 (7B)& 128   & 4096 & 5 Chunks / Turn & GRPO & 6\\
\rowcolor{R1!15} Search-AdaptR1           & Qwen2.5 (7B)& 128   & 4096 & 5 Chunks / Turn & GRPO & 6\\
 \rowcolor{Graph-R1!15} Graph-R1& Qwen2.5 (7B)               & 128   & 4096 & 5 Chunks / Turn & GRPO &3 \\
  \rowcolor{Graph-R1!15} Graph-AdaptR1& Qwen2.5 (7B)               & 128   & 4096 & 5 Chunks / Turn & GRPO &3 \\
\bottomrule
\end{tabular}}
\vspace{-1mm}
\caption{\label{tab:hyperparams}
Hyperparameter settings in Graph-R1 setting.}
\vspace{-2mm}
\end{table*}

\section{Details of Research Artifacts and Licenses}
\label{sec:artifacts}

\subsection{Datasets and Licenses}
\label{subsec:data_license}

In this work, we utilize six publicly available datasets to evaluate the multi-hop reasoning capabilities of our model. All datasets are widely used in the research community, and our use is consistent with their intended use for research and evaluation purposes.
\begin{itemize}
    \item \textbf{2WikiMultiHopQA}~\cite{2WikiMultiHopQA}: A multi-hop QA dataset using structured and unstructured data. It is distributed under the \textbf{Apache-2.0 License}.
    \item \textbf{HotpotQA}~\cite{HotpotQA}: A dataset with question-answer pairs based on Wikipedia articles, distributed under the \textbf{CC BY-SA 4.0 License}.
    \item \textbf{MuSiQue}~\cite{Musique}: A dataset for multi-hop reasoning over connected paragraphs, distributed under the \textbf{CC BY 4.0 License}.
    \item \textbf{Natural Questions (NQ)}~\cite{NQ}: A dataset consisting of queries issued to the Google search engine, distributed under the \textbf{Apache-2.0 License}.
    \item \textbf{PopQA}~\cite{PopQA}: A dataset focusing on long-tail knowledge retrieval using entity-centric questions, distributed under the \textbf{MIT License}.
    \item \textbf{TriviaQA}~\cite{TriviaQA}: A reading comprehension dataset containing question-answer-evidence triples, distributed under the \textbf{Apache-2.0 License}.
\end{itemize}

These datasets are primarily in English and are derived from public sources such as Wikipedia or Web snippets. Aligned with the experimental setup of Graph-R1~\cite{Graph-R1}, we standardize our data usage by uniformly sampling 5,120 instances for training and 128 instances for testing per dataset, thereby balancing computational workload and consistency.

\subsection{Models, Frameworks and their licenses}
\label{subsec:model_license}

We conduct our training and evaluation using the following models and frameworks:
\begin{itemize}
    \item \textbf{Language Model:} \\
    We use \textbf{Qwen2.5-7B-Instruct}~\cite{Qwen2.5} as our backbone model. Qwen2.5 is open-sourced under the \textbf{Apache-2.0 License}, allowing for research and commercial use.
    
    \item \textbf{Retrievers:} \\
    The choice of retriever depends on the specific method employed. In Search-R1, we utilize \textbf{E5}~\cite{wang2022text}. In Graph-R1, we employ hypergraph-based retrieval equipped with \textbf{bge-large-en-v1.5}~\cite{BAAIembedding}. Both embedding models are open-sourced under the \textbf{MIT License}.
    
    \item \textbf{Framework:} \\
    We implement our methods using \textbf{VeRL}~\cite{verl}, a flexible framework for reinforcement learning with LLMs. The VeRL library is open-sourced under the \textbf{Apache-2.0 License}.
\end{itemize}
\end{document}